\documentclass[Afour,sageh,times]{sagej}
\usepackage[hyphens]{url}
\usepackage{silence}
\usepackage{booktabs}
\usepackage{threeparttable}

\usepackage[colorlinks,bookmarksopen,bookmarksnumbered,citecolor=red,urlcolor=red]{hyperref}
\newcommand\BibTeX{{\rmfamily B\kern-.05em \textsc{i\kern-.025em b}\kern-.08em
T\kern-.1667em\lower.7ex\hbox{E}\kern-.125emX}}

\def\volumeyear{2016}

\usepackage[switch,pagewise]{lineno} % 'switch' puts numbers on the outer margin

\usepackage{etoolbox}

\makeatletter
\patchcmd{\@outputdblcol}{\unvbox\@outputbox}
  {\unvbox\@outputbox\par\pagebreak[0]\linenumbers}{}{}
\makeatother

\begin{document}

\sloppy
% Running head (abbreviated title and author list)
\runninghead{Singh, Hoskere, and Milillo}

% Title of your paper
\title{Multiclass Post-Earthquake Building Assessment Integrating High-Resolution Optical and SAR Satellite Imagery, Ground Motion, and Soil Data with Transformers}

% Author and affiliation setup:
\author{%
Deepank Kumar Singh\affilnum{1}, 
Vedhus Hoskere\affilnum{1,2}, 
Pietro Milillo\affilnum{1,3,4}
}

\affiliation{%
\affilnum{1}Department of Civil and Environmental Engineering, University of Houston, TX, USA\\
\affilnum{2}Department of Electrical and Computer Engineering, University of Houston, TX, USA\\
\affilnum{3}Department of Earth and Atmospheric Science, University of Houston, TX, USA\\
\affilnum{4}German Aerospace Center (DLR), Microwaves and Radar Institute, Munich, Germany
}

% Corresponding author and contact details
\corrauth{Dr. Vedhus Hoskere, 
Department of Civil and Environmental Engineering, 
University of Houston, TX, USA}

\email{vhoskere@central.uh.edu}

\begin{abstract}
Timely and accurate assessments of building damage are crucial for effective response and recovery in the aftermath of earthquakes. Conventional preliminary damage assessments (PDA) often rely on manual door-to-door inspections, which are not only time-consuming but also pose significant safety risks. To safely expedite the PDA process, researchers have studied the applicability of satellite imagery processed with heuristic and machine learning approaches. These approaches output binary or, more recently, multiclass damage states at the scale of a block or a single building. However, the current performance of such approaches limits practical applicability. To address this limitation, we introduce a metadata-enriched, transformer-based framework that combines high-resolution post-earthquake satellite imagery with building-specific metadata relevant to the seismic performance of the structure. Our model achieves state-of-the-art performance in multiclass post-earthquake damage identification for buildings from the Mw 7.8 Türkiye-Syria earthquake on February 6, 2023. Specifically, we demonstrate that incorporating metadata, such as seismic intensity indicators, soil properties, and synthetic aperture radar (SAR) damage proxy maps not only enhances the model's accuracy and ability to distinguish between damage classes, but also improves its generalizability across different regions affected by an earthquake event. Furthermore, we conducted a detailed, class-wise analysis of feature importance to understand the model’s decision-making across different levels of building damage. This analysis reveals how individual metadata features uniquely contribute to predictions for each damage class. By leveraging both satellite imagery and metadata, our proposed framework enables faster and more accurate damage assessments for precise, multiclass, building-level evaluations that can improve disaster response and accelerate recovery efforts for affected communities.

\end{abstract}


\maketitle

\section{Introduction}
In the aftermath of an earthquake, conducting a Preliminary Damage Assessment (PDA) is typically the first step in evaluating the extent of damage and determining the need for governmental assistance \cite{PreliminaryFEMA.gov}. The PDA process typically begins at the local level, where initial damage data is collected by local authorities through door-to-door inspections \cite{TurkiyeTurkiye} to determine the magnitude of damage, the impact of the disaster, and identify the worst-hit regions. These inspections can be unsafe due to falling hazards and other safety concerns due to loss of structural integrity or the risk of aftershocks.  Additionally, delays from the large number of structures to be assessed can also significantly hinder the speed of governmental aid deployment, exacerbating the recovery period for affected communities. In 2017, following the Mexico City earthquake, the Civil Engineering Association of Mexico took approximately three weeks to complete evaluations across the entire city, which delayed recovery efforts and left thousands of people without access to their homes \cite{Wang2022AVehicles,MexicansEarthquakes}. Similarly, in 2023, the Türkiye–Syria earthquakes affected 15 million people, with the preliminary damage assessments taking more than one month to complete. \cite{Erdemir2023Imaging2023,Koca2024RehabilitationTurkey,Okuyama2024EstablishmentDisasters,Tena-Colunga2021MexicoProcess,TurkiyePreventionWeb}. These cases underscore a critical need for alternative methods that can accelerate the PDA process for quicker and more effective response efforts.

The time-consuming part of the PDA process is the door-to-door building damage assessment. To help speed up access to information for decision making, researchers have proposed frameworks to automate the post-earthquake damage assessment process for the identification of the damage state at the individual building level \cite{Foroughnia2024QuantitativeData, Whitworth2022LessonsEarthquake,Voelker2024TheEarthquakes}. We discuss these frameworks by adopting a categorization based on two criteria: (i) data type, including images from  unmanned aerial vehicles (UAVs) and satellites and (ii) data processing methods, either heuristic-based approaches relying on predefined rules (e.g., \cite{Arciniegas2007Coherence-Data,Yun2015DamageCoherence,Fielding2024DamageChange,Loos2020G-DIF:Damage}) or data-driven approaches, such as machine learning (ML) and deep learning (DL) techniques, which learn patterns and make predictions based on past data and data-driven approaches (e.g., \cite{Ahmadi2023BD-SKUNet:Images,Bai2017MachineEarthquake,Joshi2021Multi-SensorFeatures,Zhang2023Earthquake-inducedClassification, Tamkuan2017FusionAssessment,Kalantar2020AssessmentImages,Xia2023AEarthquake,Varghese2023UnpairedDamage,Genova2024Vision-BasedRepair,doi:10.1177/87552930221106495}).

Researchers have employed various data sources for post-earthquake damage assessment, including aerial imagery obtained from UAVs \cite{Li2022BuildingSSD, Liu2024BDHE-Net:Data, Zhang2023Earthquake-inducedClassification}, synthetic aperture radar (SAR) satellite data \cite{Ge2020ADisasters, Macchiarulo2021City-scaleEarthquake, Bai2017MachineEarthquake, Rao2023EarthquakeLearning, Brunner2010EarthquakeImagery, Macchiarulo2024IntegratingAssessment,sarrr}, and optical satellite imagery \cite{Ahmadi2023BD-SKUNet:Images, Xia2023AEarthquake, Wang2022AImages, Joshi2021Multi-SensorFeatures, Wang2023Geometry-guidedImages, Ilmak2024DEEPEARTHQUAKES}. Data collection with UAVs can result in high-resolution and actionable imagery as demonstrated in \cite{Liu2024BDHE-Net:Data, Zhang2023Earthquake-inducedClassification}. However, scaling  UAV data collection to large areas can be time consuming and challenging due to restrictions on flying beyond visual line of sight (BVLOS) and limited battery life. For earthquakes that affect large areas, satellite imagery can be very valuable due to its wide coverage, time efficiency, and on-demand availability. In a paper by \cite{Rao2023EarthquakeLearning} the authors attempted multiclass and binary damage classification for four recent earthquakes using SAR imagery. For three out of the four earthquakes studied, the trained model achieved close to 50\% accuracy in detecting damaged buildings using binary classification. Similarly, \cite{Ahmadi2023BD-SKUNet:Images} proposed a two-stage, dual-branch UNet architecture with shared weights between the two branches using bi-temporal optical data. However, the paper highlights the problem of low accuracy for intermediate classes in multiclass classification. Additionally, a common limitation of change detection algorithms for PDA is that they rely on bi-temporal data. However, in many cases, pre-disaster images may not be readily available near the disaster date, or the built environment may have undergone significant changes since the last image was taken \cite{VITALE2024104226}. \cite{Macchiarulo2021City-scaleEarthquake} and \cite{Giardina2024CombiningMissions} utilized post-earthquake SAR imagery with a method that involved combining textural features from the imagery. The output was a map that classifies damage into five damage states at the level of city blocks, each consisting of an average of 48 buildings. However, this approach is not suitable for building-level damage assessment. A common limitation in these papers is the challenge of multiclass building-level damage assessment, primarily because nadir images make it difficult to differentiate between intermediate damage classes in earthquake-affected buildings. Additionally, the resolution of SAR imagery is often insufficient for precise building-level predictions.

Heuristic-based methods, such as the normalized difference built-up index (NDBI) derived from optical satellite imagery, utilize electromagnetic radiation to detect surface changes in objects like vegetation and buildings \cite{YuIntelligentUpdates}. While NDBI effectively highlights impervious surfaces such as buildings, it is susceptible to cloud occlusion. In contrast, SAR-derived indices like the amplitude dispersion index (ADI) \cite{Esmaeili2016ImprovedData} and damage proxy maps (DPMs) \cite{Yun2015DamageCoherence} are not affected by environmental factors \cite{SAROverview}. DPMs derived from bi-temporal SAR imagery are frequently used for rapid post-earthquake damage assessment. However, due to low resolution, DPMs are less effective at the building level, as seen in the weak correlation with building damage during the 2015 Nepal earthquake \cite{Loos2020G-DIF:Damage,Yun2015DamageCoherence}. Furthermore, damage maps produced using different satellite-based remote sensing techniques often show significant discrepancies, and extensive validation data are still required to accurately characterize the performance of these methods at both high and medium resolutions \cite{Voelker2024TheEarthquakes}. 

Data-driven approaches including machine learning (e.g., support vector machines (SVM), Random Forest (RF), etc.) and deep learning (e.g., U-net, ResNet, etc.), have widely been adopted for modern post-earthquake assessments. ML models are generally trained on tabular data. \cite{Rao2023EarthquakeLearning} proposed a ML framework that utilizes earthquake-related indices, such as peak ground acceleration (PGA), and DPMs for building-level damage assessment. \cite{YuIntelligentUpdates} further combined NDBI and ADI with DPMs and PGA to train a random forest model for PDA. The result shows ML methods outperform the traditional practice of relying solely on DPMs. However, the results are still limited to binary classification and do not perform well in multiclass classification scenarios. \cite{vbsi,xu_2022_seismic} introduced a variational causal Bayesian network to assess earthquake-induced building damage, landslides, and liquefaction using 30m bi-temporal SAR imagery, USGS shakemaps, and building footprints, but without optical data. The model outputs a 30m-resolution raster with probability estimates (ranging from 0 to 1) for each pixel. The result shows improved class discrimination compared to using only DPMs. However, the authors state that these probabilities are event-dependent e.g., both low- and high-magnitude events yield values between 0 and 1, making it difficult to compare damage severity across events and locations. The generalization capability of the approach for classifying building damage to new events is thus limited by the need for multiple thresholds to be tuned based on the ground truth acquired by other means for the event. Additionally, when multiple buildings fall within a single pixel (likely to occur in 30m resolution imagery), they all receive the same probability, making it difficult to disambiguate damage classes for a single building. Given that modern electro-optical satellite imagery is available at a 100x better resolution of 0.3m, a significant research gap in improving the resolution of assessments by an order of magnitude would be analyzing the utility of optical imagery in providing a generalizable multiclass building-level assessment without the need for manual threshold tuning. On the other hand, the training of deep learning approaches often require large amounts of high-resolution optical imagery. Recent studies \cite{Ahmadi2023BD-SKUNet:Images,Xia2023AEarthquakeb} have proposed convolutional neural network-based two-staged models for localization and classification. The results clearly demonstrate that deep neural network architectures outperform SVM and Random Forest in binary building damage classification. However, low accuracy in multiclass classification remains a challenge. For multiclass classification, Singh et al. demonstrated a significant improvement in post-hurricane building assessments using transformer vs CNN models, motivating the study of these methods for earthquakes \cite{Singh2023ClimateAssessments,Singh2023PostTransformers}.

Despite the advances described, practical implementations of post-disaster multiclass building PDA for earthquakes are challenging due to three main reasons.  First, severe class imbalance in datasets remains a critical obstacle to achieving high accuracy in multiclass classification. For example, in the widely used large-scale bi-temporal satellite imagery dataset, xBD \cite{GuptaXBD:Imagery}, which categorizes building damage into four severity levels, only 3 out of 51,500 buildings are labeled as completely destroyed. This extreme imbalance makes the training challenging and often leads to poor performance on minority classes. Second, generalization of damage assessment models across different regions and scenarios is limited as most statistical learning algorithms assume that data are independent and identically distributed (i.i.d.), meaning the training and test data come from the same underlying distribution. However, this assumption often fails in post-disaster scenarios due to regional variability, structural types, damage patterns, and limited data. Consequently, ensuring reliable model performance across diverse settings becomes challenging  \cite{quinonero2022dataset}. Third, current approaches fail to integrate diverse data sources that offer complementary information for PDA. For instance, earthquake-related indices derived from SAR provide valuable insights about variations observed on the ground surface, PGA offers local estimates of seismic intensity of the event, and optical imagery offers detailed visual information. Yet, state-of-the-art methods predominantly focus either on satellite imagery, or the metadata like PGA, DPM, etc.. This limited data fusion restricts overall classification accuracy, particularly for multiclass damage assessments. Together, these challenges highlight the need for a carefully designed PDA framework that addresses data imbalance, improves generalizability, and effectively integrates various data sources aiming to enhance multiclass PDA at building level.

\begin{figure*}
    \centering
    \includegraphics[width=1\linewidth]{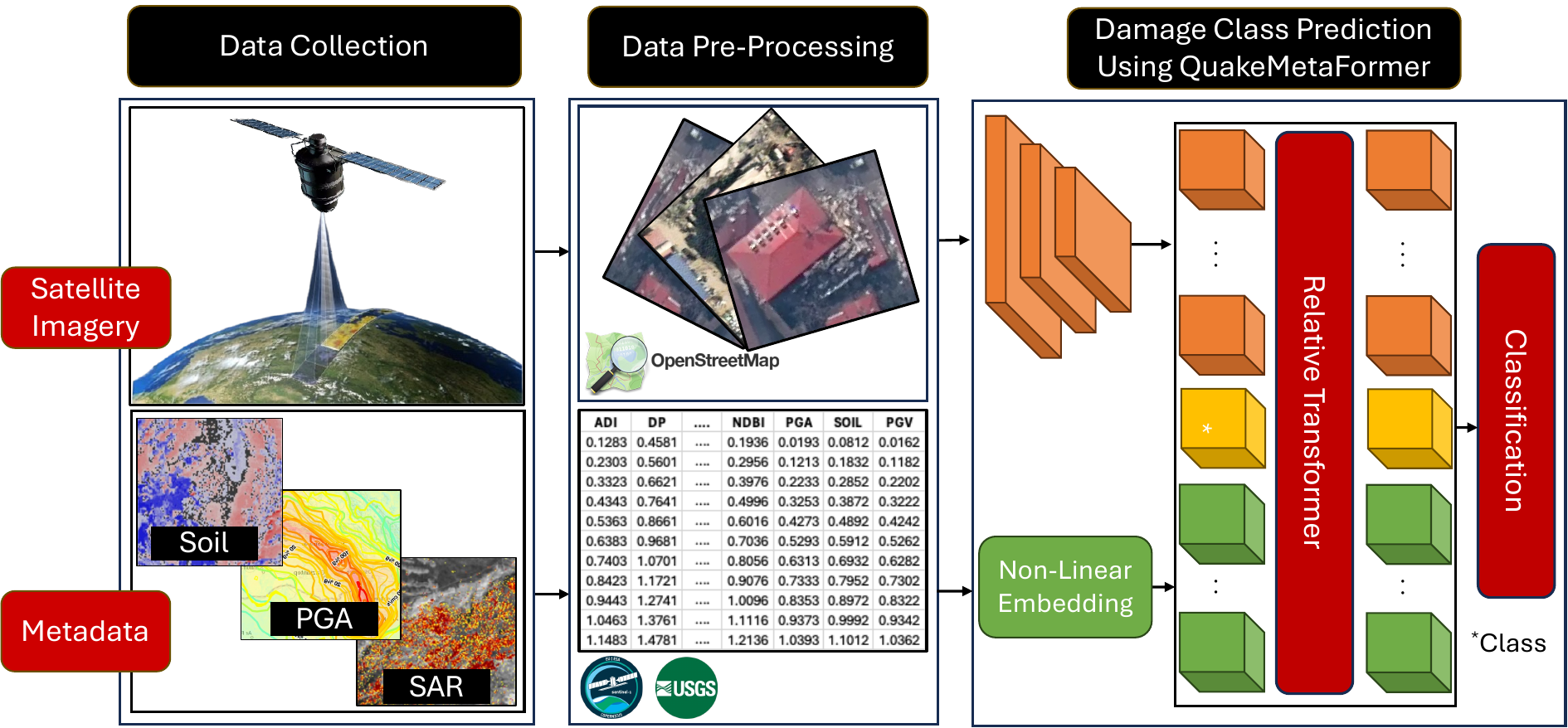}
    \caption{Proposed PDA Framework}
    \label{fig:framework}
\end{figure*}

This study proposes a novel multiclass building-level PDA framework for earthquakes (Figure \ref{fig:framework}) integrating high-resolution post-event satellite imagery with publicly available metadata with a novel transformer-based network. Metadata includes geographical attributes, disaster intensity variables, and soil properties. The novel contributions of our research include (i) the development of an earthquake-specific multiclass dataset at the building level with corresponding metadata, (ii) a transformer-based model termed QuakeMetaFormer that integrates metadata with high-resolution optical imagery for post-earthquake PDA while addressing severe class imbalance, (iii) a class-wise comparative analysis of the impact of different metadata on damage assessment, and (iv) an evaluation demonstrating the effectiveness of metadata in improving generalization across different regions affected by an earthquake. This paper is organized as follows. \textbf{Section 2} provides a detailed discussion of the proposed framework, including the processes of data collection and processing, and the architecture of QuakeMetaFormer. \textbf{Section 3} describes four experiments conducted to test our hypotheses: (i) the impact of adding metadata on model accuracy, (ii) comparison of QuakeMetaFormer and previously published machine learning frameworks, (iii) feature importance across various metadata and (iv) testing generalization capability of QuakeMetaFormer by integrating metadata. \textbf{Section 4} presents and analyzes the results from these experiments. Finally, \textbf{Section 5} concludes the paper by summarizing the key findings and their implications.

\section{Proposed Framework}
The framework outlined in Figure \ref{fig:framework} consists of three main steps. First, high-resolution satellite imagery and its corresponding metadata are collected for the earthquake-affected region, ensuring that the metadata overlaps with the geographic area covered by the satellite imagery. Next, this raw data is processed to extract building-level information, with building footprints sourced from OpenStreetMap. Finally, a novel multi-input transformer model termed QuakeMetaFormer is proposed, trained, and validated to process the extracted building-level imagery and metadata and produce a multiclass damage state. We now discuss the methodology behind these steps in more detail.

\subsection{Dataset Collection}

\begin{figure}
    \centering
    \includegraphics[width=1\linewidth]{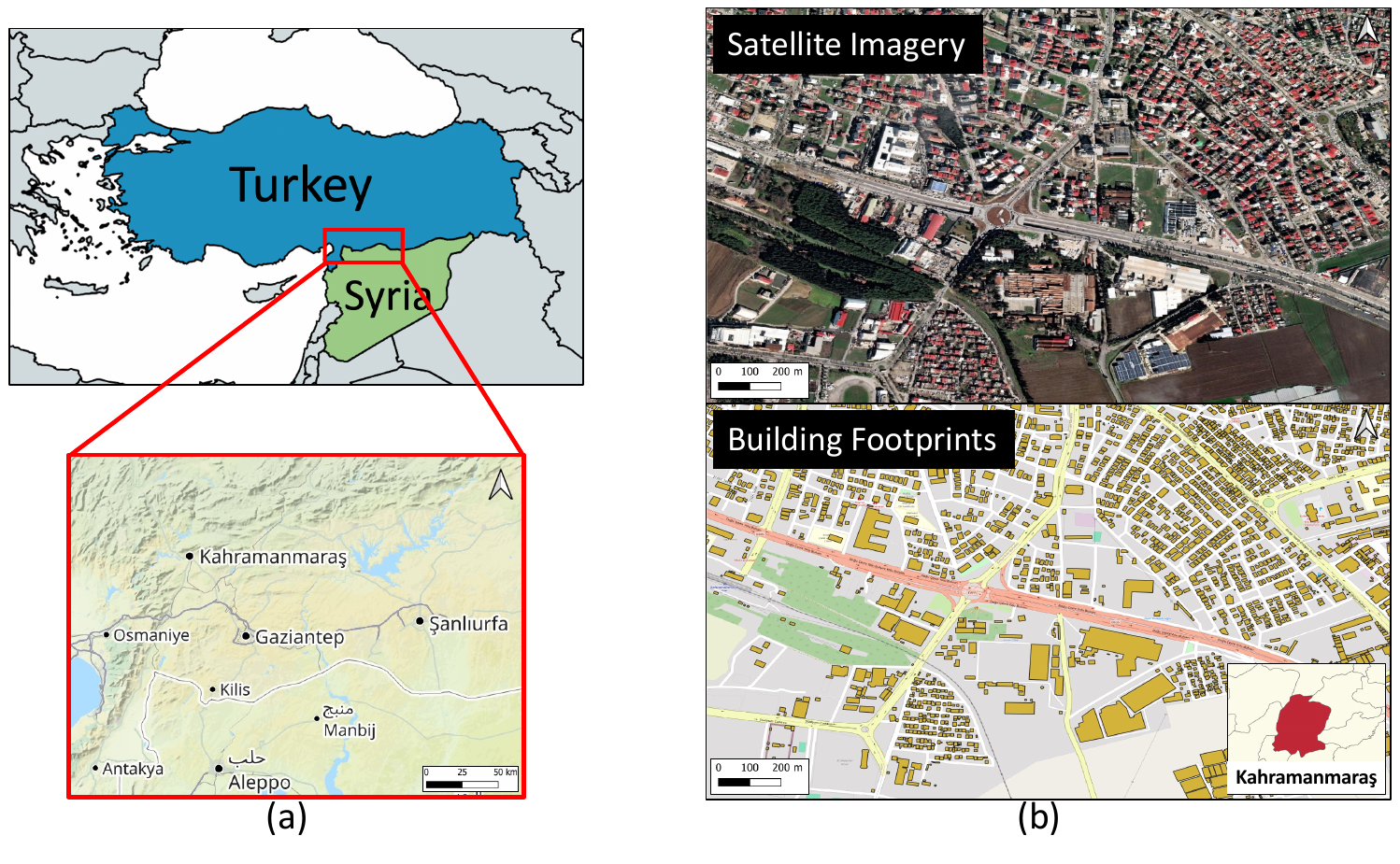}
    \caption{(a) Map showing areas of the study affected by the Türkiye-Syria earthquake, and (b) post-earthquake satellite imagery provided by MAXAR Technologies, with corresponding building footprints obtained from OpenStreetMap.}
    \label{fig:Türkiye_bf}
\end{figure}

To evaluate our proposed framework, we curate a novel dataset from buildings affected by the Türkiye-Syria earthquake of February 6, 2023, where a magnitude 7.8 quake was centered near Kahramanmaraş in southern Türkiye, as shown in Figure \ref{fig:Türkiye_bf}(a). Strong aftershocks, including those of magnitudes 7.5 and 7.6 \cite{KahramanmarasReliefWeb}, further contributed to the widespread destruction, causing additional damage. The earthquake damaged more than 200,000 buildings and led to the collapse of approximately 20,000 structures. \cite{Erdemir2023Imaging2023, Okuyama2024EstablishmentDisasters}. We now discuss details involved in the extraction of satellite imagery and corresponding metadata.

\subsubsection{Satellite Imagery}

The optical satellite imagery used in this study were obtained through MAXAR Technologies' open data program \cite{TurkeyMaxar}, which provides before and after satellite images during disasters. These images are geotagged and have a resolution of 30 cm ground sample distance (GSD) \cite{OpticalImagery}. In this study, we exclusively used post-earthquake imagery and filtered the images based on the areas affected by the earthquake. To extract building-level images, corresponding building footprints from OpenStreetMap (OSM) were utilized \cite{OpenStreetMap}. OSM is a free and open-source geographic database, that is continuously updated and maintained by a global community. Figure \ref{fig:Türkiye_bf} illustrates a sample from the dataset for the city Kahramanmaraş, highlighting the satellite imagery and corresponding building footprints. The final output for each building is a cropped section of the satellite imagery, containing three-channel (RGB) pixel-level data within the boundaries of the building footprint.

\subsubsection{Metadata}

\begin{figure}
    \centering
    \includegraphics[width=1\linewidth]{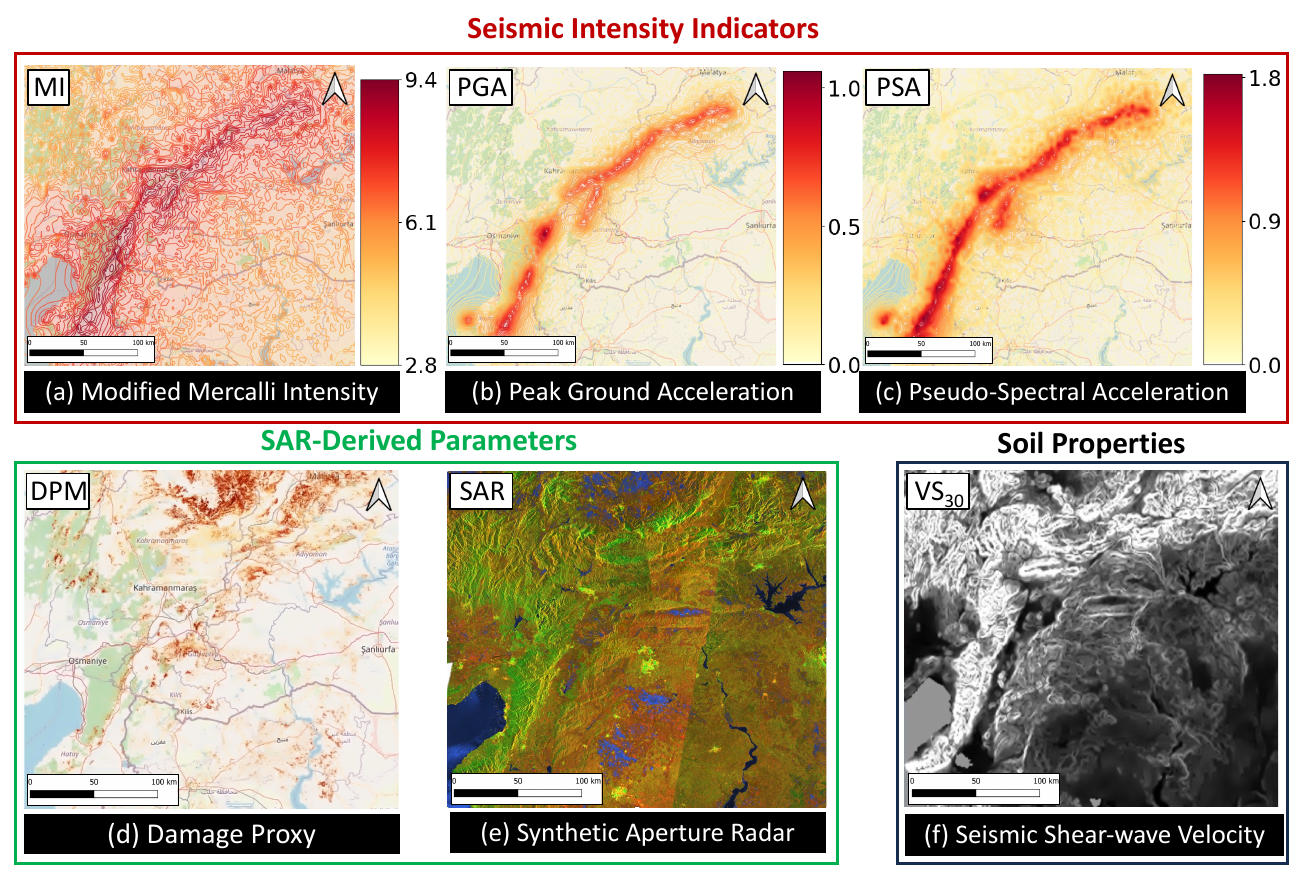}
    \caption{Metadata sources: (a)  Modified Mercalli Intensity, (b) Peak Ground Acceleration, (c) Pseudo-Spectral Acceleration, (d) Damage Proxy, (e) Synthetic Aperture Radar, and (f) Seismic Shear-wave Velocity }
    \label{fig:metadata}
\end{figure}

The different sources utilized to collect the relevant metadata for this study are shown in Figure \ref{fig:metadata}. We define metadata as data consisting of single-valued real numbers extracted using the centroid coordinates of the corresponding buildings. The metadata can be categorized into three groups: (i) seismic intensity indicators, (ii) SAR-derived parameters, and (iii) soil properties. 

Seismic intensity indicators include the Modified Mercalli Intensity (MI or MMI), Peak Ground Acceleration (PGA), Peak Ground Velocity (PGV) and Pseudo-Spectral Acceleration (PSA) at intervals of 0.3 s, 1.0 s, and 3.0 s. MI assesses earthquake impact by considering its effects on people, structures, and the landscape, with values ranging from 1 to 10, where higher numbers indicate more intense shaking and damage. PGA is a quantitative measure of the maximum ground acceleration during an earthquake, particularly relevant for short buildings (up to seven stories) \cite{EarthquakeSurvey}. PGV is an indicator of hazard for taller buildings \cite{EarthquakeSurvey}. PSA accounts for the hazard to buildings but is more closely aligned with how buildings respond to seismic forces at different frequencies, compared to peak ground motion parameters. These indices offer meaningful insights into earthquake severity and were obtained through the USGS website \cite{MSequence}. Additionally, the final condition of a structure can result not only from the primary earthquake but also from subsequent aftershocks. To account for the impact of aftershocks, we have included the top five earthquakes by intensity in our metadata for all seismic intensity indicators.

SAR-derived parameters include geocoded single look complex (SLC) amplitude data and DPMs. Geocoded SLC amplitudes  data provides information about back-scattered amplitude of the Earth's surface in the microwave wavelength. In this study, we used VV (vertical transmit, vertical receive) and VH (vertical transmit, horizontal receive) polarization values derived from Sentinel-1 ground range detected (GRD) data. These polarizations reveal details about surface orientation and scattering properties. DPMs serve as indicators of surface changes, and are derived by detecting changes in coherence between pre- and post earthquake images. These surface variations are utilized in buildings to identify structural changes, which in turn aids in damage assessment. SAR data was accessed through the Alaska Satellite Facility (ASF) via NASA's Earthdata Search API \cite{EarthdataEarthdata}, while DPMs were extracted from the Jet Propulsion Laboratory's data repository \cite{JPLShare}. 

To account for soil properties, we incorporate the average shear-wave velocity to a depth of 30 meters, known as VS30 \cite{Yong2016CompilationStates}. VS30 is widely used as a parameter to characterize site response in simplified earthquake-resistant design. VS30 maps were downloaded from the USGS website \cite{Vs30Data}.

\begin{figure*}
    \centering
    \includegraphics[width=1\linewidth]{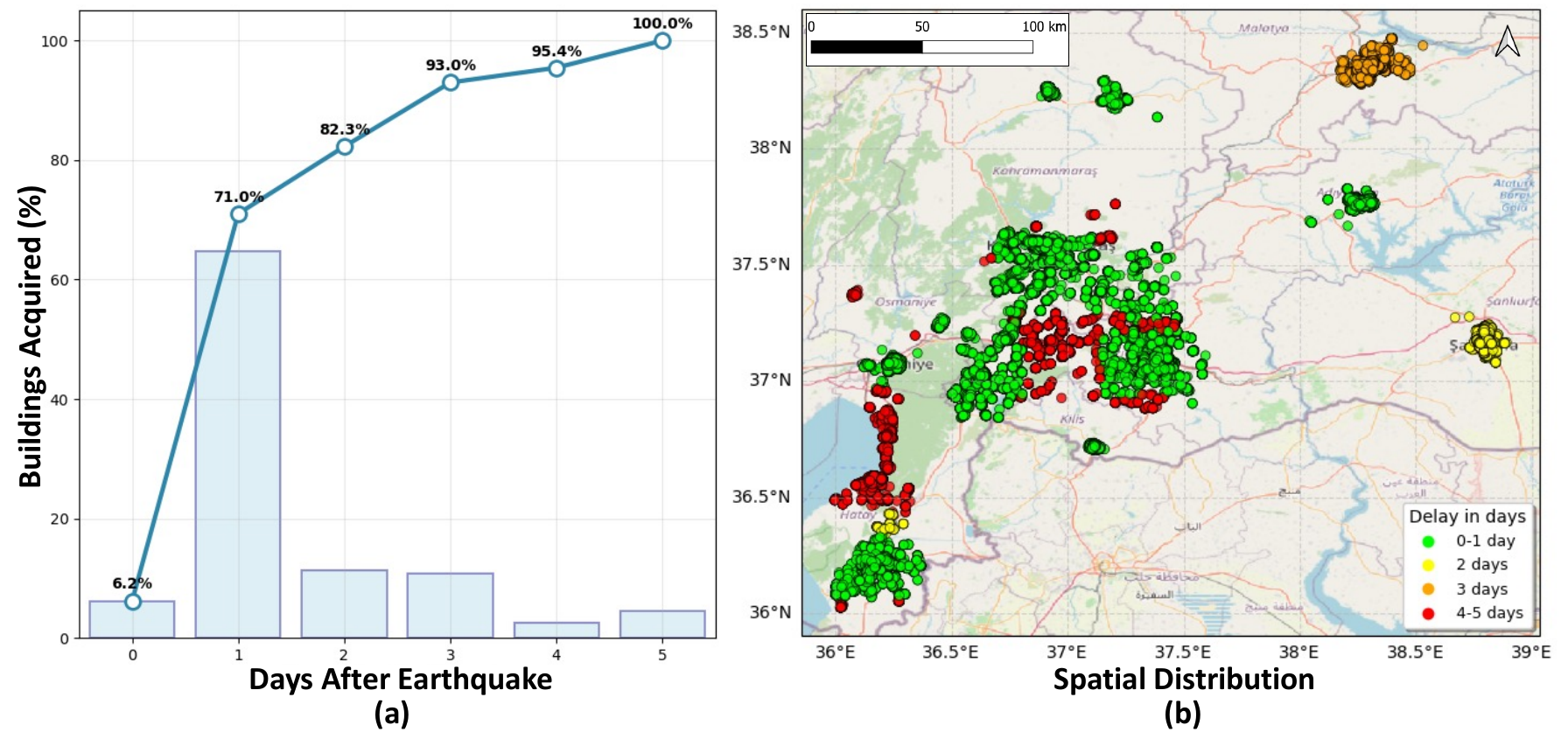}
    \caption{Optical Satellite Imagery Delays: (a) Acquisition over Time and (b) Spatial Distribution.}
    \label{fig:acquisition_plot}
\end{figure*}

Delays in data acquisition critically affects timely post‐disaster damage assessment, with acquisition timelines typically scaling with the size of area to be surveyed. Figure \ref{fig:acquisition_plot}(a) and  \ref{fig:acquisition_plot}(b) shows progress of data acquisition over time, and spatial distribution across the study area respectively. Around 71\% of the optical satellite data for the affected region were acquired within 1-day delay, reaching 100\% by day 5. Also, Figure \ref{fig:acquisition_plot}(b) suggests that data collection was prioritized for major cities near the epicenter, while remote areas and non-cities seems to experienced 3–5-day delays. The utility of the collected optical imagery is dependent on the weather conditions prevalent at the time as cloud cover may occlude buildings necessitating reacquisition. For metadata, SAR‐derived parameters (acquired Feb 11, 2023) incurred a 5-day delay, whereas USGS seismic intensity data are generally real‐time within the USA, but can be delayed 20–45 min for large earthquakes outside the US \cite{usgsdelay}.

\subsection{Data Pre-Processing }

An important step after collecting the imagery and the metadata, is to accurately map the ground truth of damage state of the building with the satellite imagery and it's corresponding metadata. The ground truth damage state for the buildings was acquired from the Turkish Ministry of Environment, Urbanization and Climate Change, Çevre, Şehircilik ve İklim Değişikliği Bakanlığı (CSB) in Turkish \cite{2023Online}. These data were processed by Gece Yazılım and Yer Çizenler respectively. The damage severity in the dataset were classified into 4 damage classes, numbered 1-4, Slightly Damaged, Heavily Damaged, Needs to be demolished, and Collapsed. Mapping the ground truth with the dataset was done in two steps: mapping the ground truth to the building footprint, and then retrieving its corresponding meta data. In the first step, the nearest building footprint for each ground truth damage location (latitude, longitude) was identified by calculating spatial distances using an R-tree index \cite{GuttmanR-TREES.SEARCHING}. Then, the data was filtered based on whether each ground truth damage location fell within its corresponding building footprint to ensure accurate mapping. In the second step, metadata was associated with the mapped building footprints by using the geometric centroid of each footprint as a query point. For contour shapefile data, such as seismic intensity indicators, interpolation was employed to estimate values at the query point. SAR-derived parameters and soil properties were extracted from raster GeoTIFF files. The raster data, originally in a global coordinate system measured in degrees (EPSG:4326), was reprojected into a local coordinate system measured in meters (EPSG:32637) to ensure precise alignment with the ground truth damage locations. This approach ensured that each building point was accurately associated with the relevant metadata. 
\begin{figure*}
    \centering
    \includegraphics[width=1\linewidth]{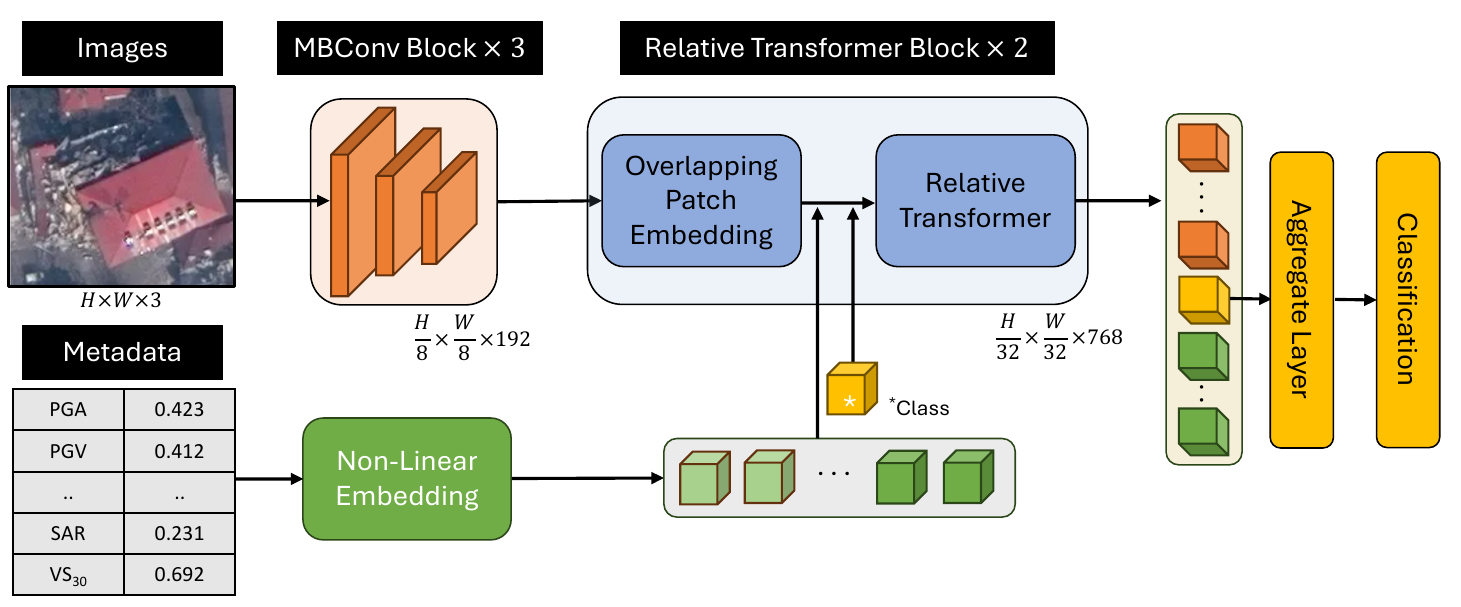}
    \caption{MetaFormer Architecture: Initial stages use convolutional downsampling, followed by relative transformer layers for fusing image and metadata}
    \label{fig:architecture}
\end{figure*}
In addition to our primary dataset, we used another dataset developed by \cite{YuIntelligentUpdates}, including the ground truth labels and metadata. We then extracted the corresponding satellite imagery using our methodology and mapped it to the metadata for model training. This dataset includes five metadata features: damage proxy (DP) derived from Sentinel-1 and ALOS-2 PALSAR-2, ADI, NDBI, and PGA. The dataset has 5 damage classes numbered 0-4, No damage, Slightly Damaged, Heavily Damaged, Needs to be demolished, and Collapsed. The details of both datasets are summarized in the table \ref{tab:dataset_summary}.

\begin{table}[h]
\small\sf\centering
\caption{Summary of Datasets}
\resizebox{0.5\textwidth}{!}{%
\begin{tabular}{l l l}
\toprule
\textbf{Name}        & \textbf{Türkiye-L}                     & \textbf{Türkiye-S}   \\ 
\midrule
\textbf{MetaData Source}           & USGS, ASF, JPL (Our study)                                   & \cite{YuIntelligentUpdates}                            \\ 
\textbf{Imagery Source}         &    Maxar (Our study)                 & Maxar (Our study)\\  
\textbf{Dataset Size}            & 81,424 (Train) + 14,369 (Test)               & 13,277 (Train) + 2,343 (Test)             \\ 
\textbf{No. of Classes}   & 4 Classes (1 to 4)                                   & 5 Classes (0 to 4)                          \\ 
\textbf{Region}           & Türkiye (All major cities)                    & Kahramanmaraş City \\

\midrule
\end{tabular}}

\label{tab:dataset_summary}
\end{table}

\subsection{QuakeMetaFormer}

Leveraging the MetaFormer architecture proposed by \cite{DiaoMetaFormerRecognition},  illustrated in Figure \ref{fig:architecture}, QuakeMetaFormer is a multi-input model that takes both image data and metadata as inputs optimized for post-earthquake PDA by accounting for extreme class imbalance. MetaFormer employs a hybrid approach where convolution is utilized to encode visual data, while transformer layers focus on combining this visual information with metadata, which is encoded using a non-linear embedding. The initial three stages of MetaFormer primarily utilize MBConv blocks, followed by the adoption of Relative Transformer blocks in the final two stages. To optimize computational efficiency, overlapping patch embedding is implemented for tokenizing the feature map and performing downsampling. The model configuration utilizes key hyperparameters listed in Table \ref{tab:hyperparameters}. The hyperparameters from the original MetaFormer paper were directly adopted from \cite{DiaoMetaFormerRecognition}.

QuakeMetaFormer modifies MetaFormer to handle the class-imbalanced nature of datasets typically available for post-earthquake damage classification where examples of severe damage are naturally fewer compared to buildings with little or no damage. Specifically, we implement a modified focal loss that incorporates the class weights into the focal loss introduced by \cite{Lin2020FocalDetection}. Class weights are scaling factors to give more importance to underrepresented classes, calculated using the inverse frequency of each class in the dataset. The modified focal loss is defined as:
\begin{equation}
    \text{Modified Focal Loss} = w_t \cdot (1 - p_t)^\gamma \cdot \text{CE}
\end{equation}
where \(w_t\) represents the class weight for the true class, \(p_t\) is the predicted probability of the true class, \(\gamma\) is the focusing parameter, and \(\text{CE} = -\log(p_t)\) denotes the cross-entropy loss. We found that a gamma value of 4 worked best for our dataset, effectively focusing on hard-to-classify samples (see Table \ref{tab:embedding}).

\begin{table}
\centering
\caption{Model Hyperparameters}
\begin{tabular}{l c} % Alignment options
\toprule
\textbf{Parameter}              & \textbf{Value}       \\ 
\midrule
Batch Size             & 20                  \\ 
Image Size             & 384                 \\ 
Optimizer              & AdamW               \\ 
Base Learning Rate     & 0.00005             \\ 
Weight Decay           & 0.05                \\ 
Learning Rate Scheduler & Cosine              \\ 
Epochs                 & 150                 \\ 
\bottomrule
\end{tabular}
\label{tab:hyperparameters}
\end{table}

\begin{table*}
\centering
\caption{Comparison of Various Metadata Embeddings, Loss Functions, and F1 Scores}
\begin{tabular}{l l l c l}
\toprule
\textbf{Dataset} & \textbf{Embedding} & \textbf{Loss Function} & \textbf{F1 Score} & \textbf{Description} \\
\midrule
Türkiye-L & MLP              & MFL & \textbf{0.49} & Standard feed-forward stack \\
Türkiye-L & MLP          & CEL & {0.31}          & Standard feed-forward stack\\
Türkiye-L & ResNorm          & MFL & 0.48         & Residual connections with LayerNorm \\
Türkiye-L & DeepResNorm      & MFL & 0.48         & Stacked ResNorm blocks \\
Türkiye-L & DeepMLP          & MFL & 0.47         & Extended feed-forward network \\
Türkiye-L & Transformer      & MFL & 0.43         & Standard Transformer encoder \\
Türkiye-L & LargeTransformer & MFL & 0.43         & 2-layer Transformer encoder \\
\midrule
Türkiye-S & MLP              & MFL & \textbf{0.56} & Standard feed-forward stack \\
Türkiye-S & MLP          & CEL & {0.34}          & Standard feed-forward stack\\
Türkiye-S & ResNorm          & MFL & 0.54         & Residual connections with LayerNorm \\
Türkiye-S & DeepMLP          & MFL & 0.54         & Extended feed-forward network \\
Türkiye-S & DeepResNorm      & MFL & 0.53         & Stacked ResNorm blocks \\
Türkiye-S & Transformer      & MFL & 0.32         & Standard Transformer encoder \\
Türkiye-S & LargeTransformer & MFL & 0.23         & 2-layer Transformer encoder \\
\bottomrule
\end{tabular}
\label{tab:embedding}

\vspace{0.5em}
\raggedright
\footnotesize{CEL: Cross-Entropy Loss, MFL: Modified Focal Loss}

\end{table*}

Another modification we introduced focuses on enhancing the incorporation of metadata through different embedding strategies. We compared multiple approaches: ResNorm, MLP, and Transformers. ResNorm combines residual connections with layer normalization, applying linear transformations and ReLU activations in a skip-connected fashion. In contrast, the MLP approach is a feed-forward stack of linear layers and activations followed by a final layer normalization. Meanwhile, the Transformer-based model leverages multi-head self-attention and feed-forward sublayers to encode metadata embeddings. We implemented two versions of each of these embeddings with deeper architectures. The Table \ref{tab:embedding} below presents the F1 scores for each dataset, along with brief explanations of the modifications made in the deeper variants.The results show that the MLP embedding strategy consistently achieves the highest F1 scores for both datasets, indicating that the feed-forward architecture performs robustly compared to the other methods. Additionally, while the deeper versions of ResNorm and MLP exhibit comparable performance to their standard counterparts, the Transformer-based approaches record noticeably lower F1 scores, particularly on the Türkiye-S dataset.

In addition to the QuakeMetaFormer model, we also employ a variant that only takes imagery as input and excludes the metadata embedding segment for metadata. This variant, referred to as the QuakeImageFormer in this study, serves as a baseline for evaluating the impact of adding metadata, while keeping the rest of the model architecture the same for a fair comparison.

\section{Experiments}
In this section, we evaluate the performance of the proposed PDA framework by conducting four experiments: (i) impact of integrating metadata with satellite imagery on the overall model performance, (ii) comparing the performance of the proposed framework with two previously published state-of-the-art machine learning frameworks, (iii) assessing the influence of individual metadata features on damage state prediction, and (iv) evaluating the effect of metadata in improving the trained model's generalizability to different regions. 

\subsection{Impact of Adding Metadata on Performance: QuakeMetaFormer}

Satellite imagery provides essential visual information for damage assessment, while earthquake-related metadata, such as PGA, SAR, VS30, etc., offers unique insights into event intensity and structural changes. However, relying on either data source individually often lacks the detail necessary for distinguishing between intermediate damage classes. The objective of this experiment is to evaluate how incorporating metadata with satellite imagery impacts the performance of the QuakeMetaFormer model. To achieve this, we conducted three experiments: (i) QIF-Türkiye-L, where the QuakeImageFormer (QIF) was trained using only satellite imagery to establish baseline performance, (ii) QMF-Türkiye-L, where the QuakeMetaFormer (QMF) model was trained using both satellite imagery and all the metadata, including aftershock seismic intensity indicators; and (iii) QMF-Türkiye-L1, where the QuakeMetaFormer model was trained using both satellite imagery and metadata, but with the seismic intensity indicators for aftershocks removed, retaining only data from the main event of the largest magnitude.  Similarly, two additional experiments, QIF-Türkiye-S and QMF-Türkiye-S, were conducted on the Türkiye-S dataset. Comparing the results from these experiments allows us to understand the impact of adding metadata to satellite imagery in multiclass damage classification. Additionally, to evaluate the impact of metadata on the model's ability to interpret imagery, we conducted four additional experiments. In these experiments, metadata was incorporated during training but masked during testing. Three experiments, QMF-Türkiye-L-M, QMF-Türkiye-L1-M and QMF-Türkiye-S-M, simulate scenarios where all metadata are masked during testing, while QMF-Türkiye-L1-M-SAR represents the case where only SAR-derived parameters (SAR-VV, SAR-VH, and DPM) are masked. This setup is particularly relevant for rapid post-earthquake damage assessment, where high-latency SAR data may be delayed. For example, \cite{VITALE2024104226} notes that for high-resolution SAR acquisitions in spotlight mode, coverage latency can extend up to 11 days for areas larger than 9,247 km², making it impractical for timely response over large disaster zones such as the ~350,000 km² region affected by the 2023 Turkey–Syria earthquake. These experiments allow us to evaluate the practical applicability of QuakeMetaFormer when certain metadata are unavailable at inference time.

Different models trained are summarized in Table \ref{tab:experiment_summary}. In each of our experiments, we utilize 85\% of the data for training and reserve 15\% for the testing set to evaluate model performance. Each experiment is discussed in detail in the following sections.

\begin{table*}
  \centering
  \begin{threeparttable}
    \caption{Summary Table: Impact of Adding Metadata}
    \label{tab:experiment_summary}
    \begin{tabular}{l c c c c}
      \toprule
      \textbf{Experiment Name}   & \textbf{Architecture}      & \textbf{Data}               & \textbf{Metadata} & \textbf{Metadata-Masked} \\ 
      \midrule
      QMF-Türkiye-L       & QuakeMetaFormer (QMF)      & Türkiye-L    & Yes         & No   \\ 
      QIF-Türkiye-L       & QuakeImageFormer (QIF)     & Türkiye-L    & No          & N/A  \\ 
      QMF-Türkiye-L-M     & QuakeMetaFormer (QMF)      & Türkiye-L    & Yes         & Yes  \\
      QIF-Türkiye-L-KAH$^*$  & QuakeImageFormer (QIF)     & Türkiye-L    & No          & N/A  \\ 
      QMF-Türkiye-L1      & QuakeMetaFormer (QMF)      & Türkiye-L    & Yes         & No   \\ 
      QMF-Türkiye-L1-M    & QuakeMetaFormer (QMF)      & Türkiye-L    & Yes         & Yes  \\ 
      QMF-Türkiye-L1-M-SAR& QuakeMetaFormer (QMF)      & Türkiye-L    & Yes         & Yes  \\
      \midrule
      QMF-Türkiye-S       & QuakeMetaFormer (QMF)      & Türkiye-S    & Yes         & No   \\ 
      QMF-Türkiye-S-M     & QuakeMetaFormer (QMF)      & Türkiye-S    & Yes         & Yes  \\ 
      QIF-Türkiye-S       & QuakeImageFormer (QIF)     & Türkiye-S    & No          & N/A  \\ 
      \bottomrule
    \end{tabular}
    \begin{tablenotes}[flushleft]
      \footnotesize
      \item[*] Tested only on Kahramanmaraş Region
    \end{tablenotes}
  \end{threeparttable}
\end{table*}

\subsection{Comparison of QuakeMetaFormer and Machine Learning Frameworks}

To evaluate the performance of our proposed framework against other state-of-the-art PDA frameworks for earthquake damage assessment, we compared different architectures trained on the same dataset. For this experiment, we trained the QuakeMetaFormer model (QMF-Türkiye-S) on the Türkiye-S dataset (Imagery and metadata) and compared its performance with the best-performing machine learning model (random forest) reported by \cite{YuIntelligentUpdates}, which was trained on the same metadata from the Türkiye-S dataset. On scanning the literature, we found another paper, \cite{Rao2023EarthquakeLearning} that conducted a multiclass classification study for post-earthquake damage evaluation across regions impacted by 2015 Gorkha, 2017 Puebla, 2020 Puerto Rico, and 2020 Zagreb earthquakes. However, the datasets from \cite{Rao2023EarthquakeLearning} were not accessible, so we report the average performance of our model across both our datasets and the average performance across all datasets in Rao et al. The experiments are summarized in Table \ref{tab:experiment_summary_2}.
\begin{table}
\centering
\caption{Summary Table: QMF vs ML Frameworks}
\resizebox{0.5\textwidth}{!}{%
  \begin{tabular}{l c c c c}
    \toprule
    \textbf{Experiment Name}   & \textbf{Architecture}       & \textbf{Source}            \\ 
    \midrule
    QMF-Türkiye-S       & QuakeMetaFormer (QMF)               & Our Study     \\ 
    RF-Yu       & Random Forest (RF)&    \cite{YuIntelligentUpdates}          \\
    RF-Rao & Random Forest (RF) &  \cite{Rao2023EarthquakeLearning} \\
    \bottomrule
  \end{tabular}%
}
\label{tab:experiment_summary_2}
\end{table}

\subsection{Feature Importance Across Metadata}

Various metadata provide different insights related to earthquake damage assessment. To assess the contribution of individual metadata features, we  conducted a detailed analysis using SHapley Additive exPlanations (SHAP) \cite{Lundberg2017APredictions}. SHAP provides a robust method for understanding the output of machine learning models by perturbing each feature and measuring the resulting changes in prediction probabilities. These changes are then used to assign an importance score based on the feature's contribution to the predictions and reveal how specific features increase the model’s confidence (probabilities) in predicting certain damage classes. For this experiment, we applied SHAP to the QMF-Türkiye-L1 model to evaluate the influence of various metadata inputs on damage state prediction. This analysis helps in understanding the impact of different earthquake-related indices on model performance. We also performed a class-wise analysis of metadata features to better understand the impact on different classes.

\subsection{Generalization Across Regions}

\begin{figure}
    \centering
    \includegraphics[width=1\linewidth]{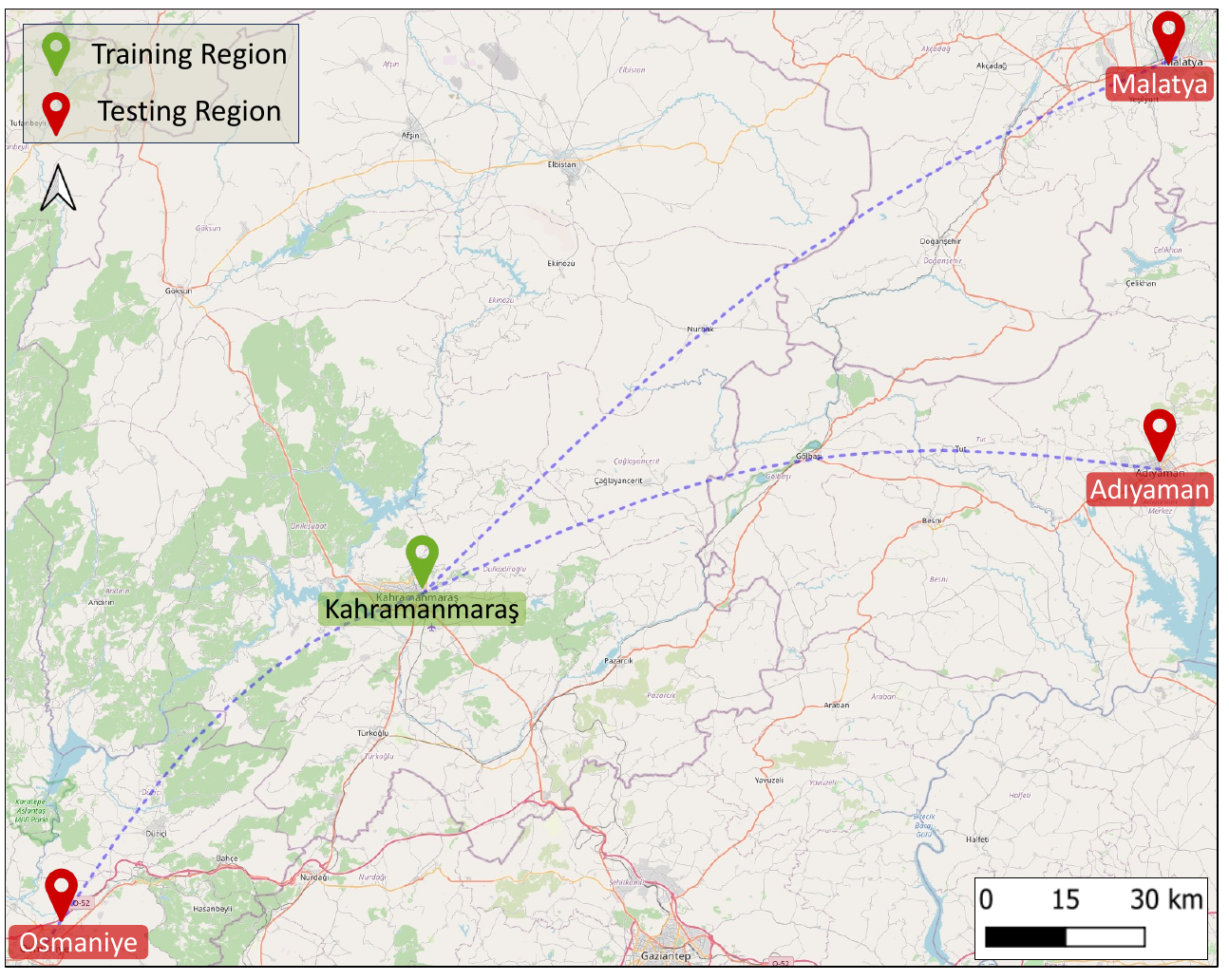}
    \caption{Map of Training and Testing Regions}
    \label{fig:cross-regional map}
\end{figure}

To evaluate the generalization capability of the QuakeMetaFormer model, we divided the Türkiye-L dataset into four major regions: Kahramanmaraş, Adıyaman, Osmaniye, and Malatya. The locations of the training and testing regions are illustrated in Figure \ref{fig:cross-regional map}. For this experiment, we trained two separate models on the Kahramanmaraş region (i) QuakeMetaFormer with metadata and imagery, and (ii) QuakeImageFormer with imagery only. These two trained models were then tested on the remaining three regions (see Table \ref{tab:generalization_experiments}). As discussed earlier, the similarity in data distribution between training and testing regions greatly impacts a model’s generalization capabilities, models tend to generalize better when the test data closely matches the training distribution \cite{quinonero2022dataset}. This experiment aims to test our hypothesis that incorporating metadata can enhance generalization, particularly by providing additional information that may help bridge distribution gaps across various regions.

\begin{table*}
\centering
\caption{Generalization Across Regions Experiments}

\begin{tabular}{l c c c}
\toprule
\textbf{Experiment Name}     & \textbf{Region}   & \textbf{Architecture} & \textbf{Experiment Setup} \\ 
\midrule
KAH-QMF-KAH             & Kahramanmaraş              & QuakeMetaFormer & Metadata + Imagery \\ 
KAH-QIF-KAH           & Kahramanmaraş              & QuakeImageFormer & Imagery Only  \\ 
KAH-QMF-ADİ       & Adıyaman                   & QuakeMetaFormer & Testing    \\ 
KAH-QIF-ADİ      & Adıyaman                   & QuakeImageFormer & Testing    \\ 
KAH-QMF-OSM       & Osmaniye                   & QuakeMetaFormer & Testing    \\ 
KAH-QIF-OSM      & Osmaniye                   & QuakeImageFormer & Testing    \\ 
KAH-QMF-MAL       & Malatya                  & QuakeMetaFormer & Testing    \\ 
KAH-QIF-MAL      & Malatya                  & QuakeImageFormer & Testing    \\ 
\bottomrule
\end{tabular}
\label{tab:generalization_experiments}
\end{table*}

\subsection{Classification Metrics}

To evaluate the experimental results, we employed five standard classification metrics, including accuracy, precision, recall, F1 score, and the average area under the receiver operating characteristic (AUC-ROC) curve. Accuracy is the overall correctness of the model, calculated as the ratio of correctly predicted instances to the total number of instances. Precision measures the model’s ability to minimize false positives by calculating the proportion of correctly predicted positive instances, averaged across all classes to ensure balanced consideration. Similarly, recall assesses the model’s effectiveness in identifying true positives by averaging recall values across classes. The F1 score, combines precision and recall, serves as a balanced metric for accuracy, particularly in datasets with imbalanced class distribution. The AUC-ROC curve, widely used as a performance evaluation metric in classification models, particularly in fields like remote sensing and natural hazard monitoring, provides a comprehensive assessment of a classifier's ability to distinguish between classes \cite{Alatorre2011IdentificationImagery,ChangLandslideModels,YuIntelligentUpdates,Fawcett2006}. The AUC-ROC curve offers a comprehensive view of the classifier’s ability to distinguish between positive and negative instances by plotting the true positive rate (TPR) against the false positive rate (FPR). A higher AUC-ROC score indicates better ability to separate positive and negative instances.\cite{Alatorre2011IdentificationImagery,ChangLandslideModels,YuIntelligentUpdates}. Table \ref{tab:metrics} summarizes the metrics and their formulas.

\begin{table}
\centering
\caption{Classification Metrics}
\begin{tabular}{lc}
\toprule
\textbf{Metric}            & \textbf{Formula} \\ 
\midrule
Accuracy            & $\frac{\text{TP + TN}}{\text{TP + TN + FP + FN}}$ \\[8pt]
Precision           & $\frac{1}{N} \sum_{i=1}^{N} \frac{\text{TP}_i}{\text{TP}_i + \text{FP}_i}$ \\[8pt]
Recall              & $\frac{1}{N} \sum_{i=1}^{N} \frac{\text{TP}_i}{\text{TP}_i + \text{FN}_i}$ \\[8pt]
F1 Score            & $\frac{1}{N} \sum_{i=1}^{N} 2 \times \frac{\text{Precision}_i \times \text{Recall}_i}{\text{Precision}_i + \text{Recall}_i}$ \\[8pt]
AUC-ROC                    & $\frac{1}{N} \sum_{i=1}^{N} \text{AUC-ROC}_i$ \\ 
\bottomrule
\end{tabular}
\label{tab:metrics}
\end{table}

\section{Results and Discussion}

This section provides a detailed analysis of the four experiments discussed in the previous section, highlighting the performance outcomes and insights derived from each.

\subsection{Impact of Adding Metadata on Performance: QuakeMetaFormer}

The objective of this experiment is to assess the impact of integrating metadata with satellite imagery in improving the performance of the PDA framework. The results, as shown in Table \ref{tab:performance_comparison}, demonstrate that incorporating metadata enhances model performance across all metrics.

\begin{figure}
    \centering
    \includegraphics[width=1\linewidth]{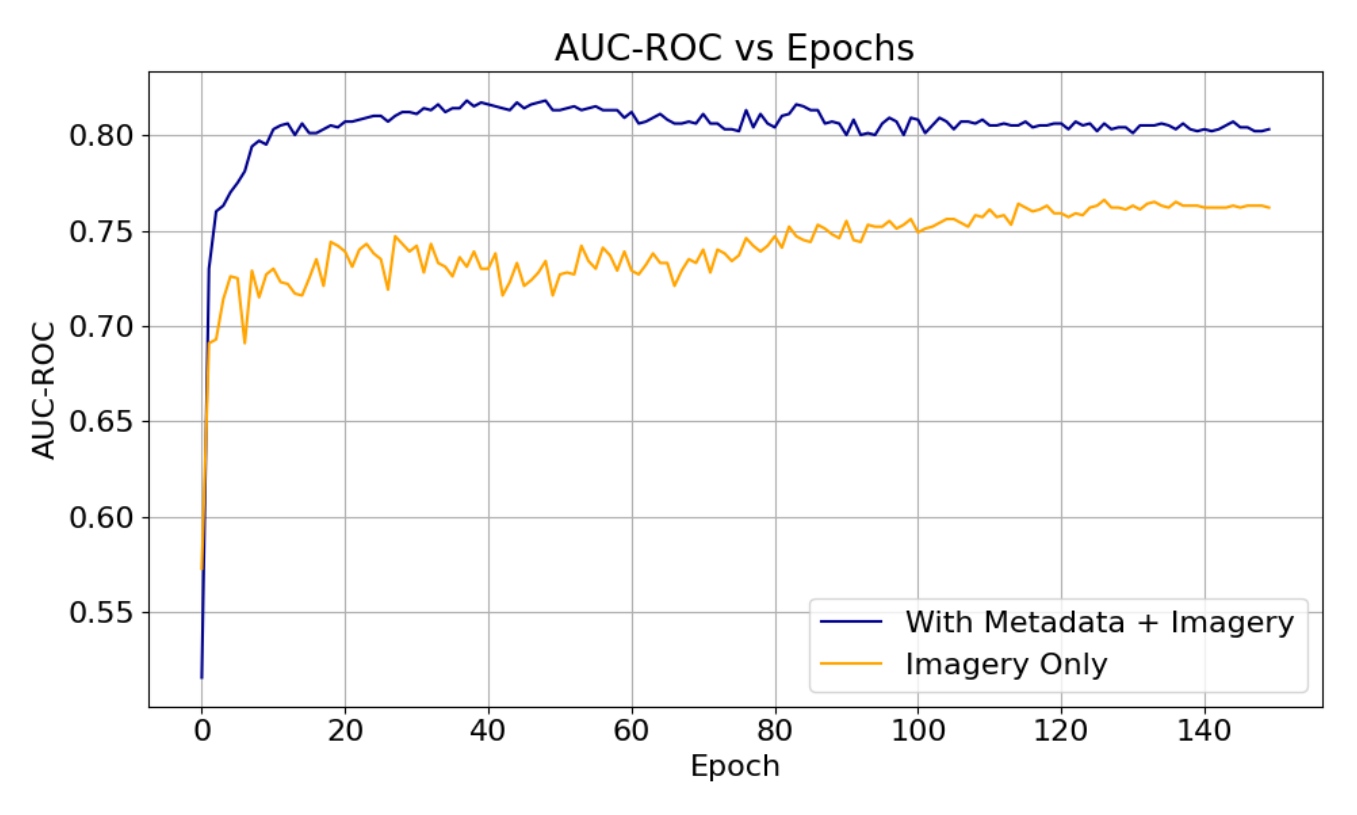}
    \caption{AUC-ROC Learning Curve}
    \label{fig:AUC-ROC Learning Curve}
\end{figure}

\begin{table*}
\centering
\caption{Performance Comparison of Different Experiments}
\label{tab:performance_comparison}
\begin{tabular}{l c c c c c}
\toprule
\textbf{Experiment}   & \textbf{Accuracy}  & \textbf{Precision} & \textbf{Recall} & \textbf{F1-Score} & \textbf{AUCROC} \\ 
\midrule
QMF-Türkiye-L            & 0.77               & 0.58               & 0.47            & 0.49              & 0.82            \\ 
QIF-Türkiye-L           & 0.70               & 0.48               & 0.43            & 0.45              & 0.76            \\
QMF-Türkiye-L-M          & 0.75               & 0.55               & 0.41            & 0.45              & 0.79            \\ 
QMF-Türkiye-L-KAH          & 0.79               & 0.61               & 0.48            & 0.50              & 0.83            \\
QMF-Türkiye-L1            & 0.76               & 0.53               & 0.44            & 0.47              & 0.81            \\ 
QMF-Türkiye-L1-M          & 0.75               & 0.52               & 0.43            & 0.46              & 0.78            \\ 
QMF-Türkiye-L1-M-SAR          & 0.76               & 0.53               & 0.43            & 0.46              & 0.80            \\
\midrule
QMF-Türkiye-S            & 0.77               & 0.74               & 0.49            & 0.56              & 0.85            \\ 
QIF-Türkiye-S           & 0.71               & 0.68               & 0.46            & 0.51              & 0.79            \\ 
QMF-Türkiye-S-M          & 0.75               & 0.70               & 0.46            & 0.53              & 0.81            \\  

\bottomrule 
\end{tabular}

\end{table*}
QMF-Türkiye-S achieved 77\% accuracy compared to 71\% for the QIF-Türkiye-S model, while for the Türkiye-L dataset, accuracy increased from 70\% to 77\%. In both sets of experiments, adding metadata resulted in a 7\% accuracy improvement. Beyond accuracy, AUC-ROC and F1-score offer a clearer evaluation of model performance, especially in imbalanced datasets where some damage classes are underrepresented. For instance, the QMF-Türkiye-S model achieved an AUC-ROC of 0.85, compared to 0.79 for the QIF-Türkiye-S model, and an F1-score of 0.56, compared to 0.51 for QIF-Türkiye-S. These improvements indicate that incorporating metadata significantly enhances the model’s ability to differentiate between multiple damage classes, providing a more reliable performance across all damage categories. The AUC-ROC learning curve (see Figure \ref{fig:AUC-ROC Learning Curve}) for Türkiye-L dataset shows the consistent improvement in performance for the models trained with metadata and without metadata for 150 epochs. QuakeMetaFormer reaches a 6\% higher AUC-ROC than the QIF model, underscoring the importance of metadata in enhancing the model's capacity for classifying damage more accurately. Another important observation is that adding metadata improves precision by 10\% in the case of Türkiye-L and 6\% in the case of Türkiye-S, demonstrating that incorporating metadata helps the model reduce false positives. This means the model becomes more reliable in correctly identifying buildings with actual damage, minimizing the number of undamaged buildings mistakenly classified as damaged. Notably, adding metadata not only improves the accuracy but also enhances the model’s ability to interpret imagery. For instance, QMF-Türkiye-L1 achieves an AUC-ROC of 0.81, compared to QMF-Türkiye-L1-M’s 0.78, reflecting only a minimal decrease in performance. Similarly, QMF-Türkiye-L1-M-SAR, which simulates a scenario where SAR derived data is masked during testing, achieves an AUC-ROC of 0.80. The drop in AUC-ROC is minor compared to the case of QIF-Türkiye-L1, where no metadata is added during training. This experiment underscores the model’s applicability and robustness in practical scenarios where where high-latency data, such as SAR derived parameters (SAR-VV, SAR-VH, and DPM) may be delayed.

\begin{figure}
    \centering
    \includegraphics[width=1\linewidth]{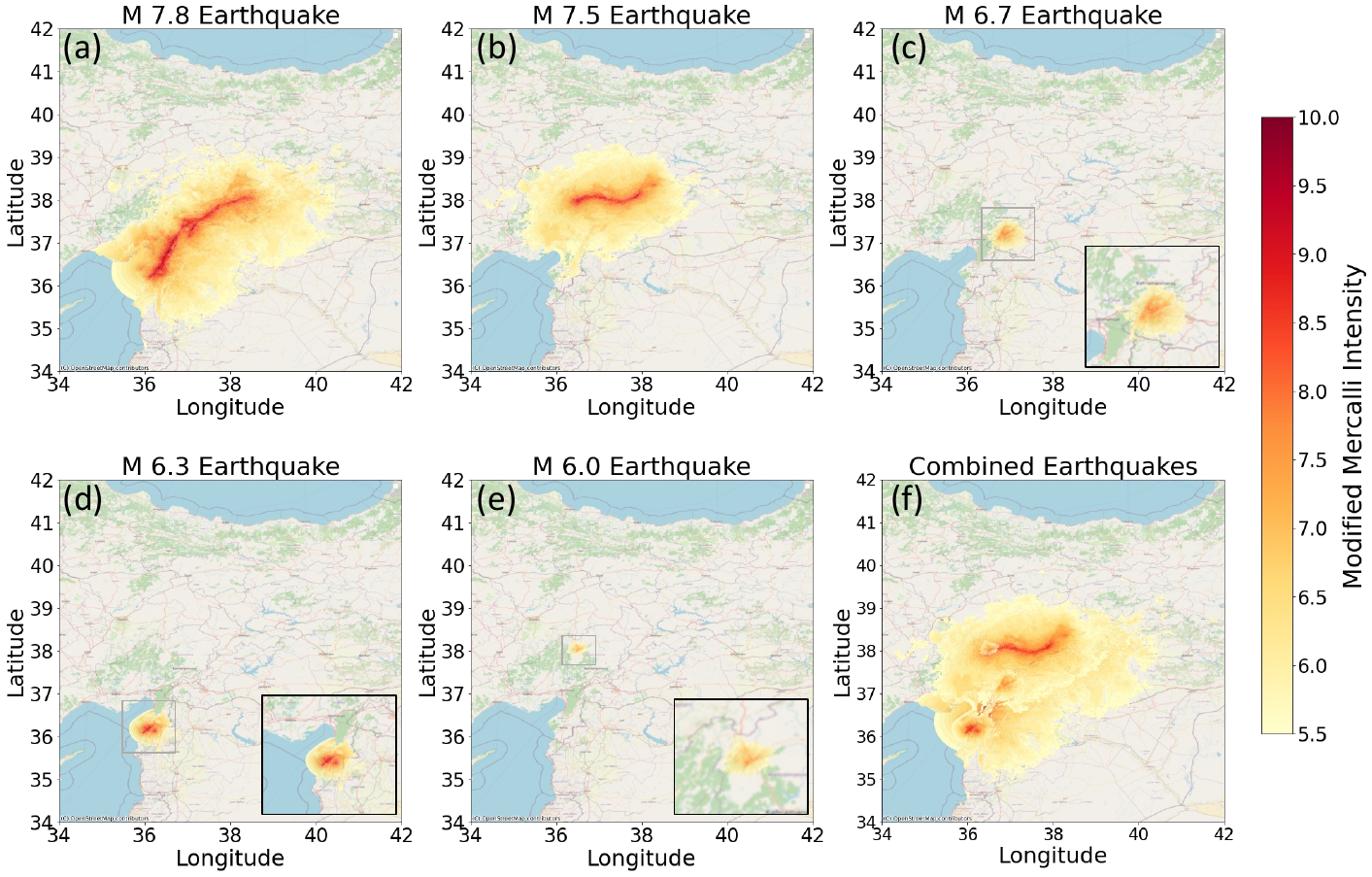}
    \caption{Modified Mercalli Intensity heatmaps for top 5 earthquakes: (a) M 7.8, (b) M 7.5, (c) M 6.7, (d) M 6.3, (e) M 6.0, and (f) Combined}
    \label{fig:multiEQ}
\end{figure}

We evaluated the effect of incorporating aftershock metadata (seismic intensity indicators) by comparing QMF-Türkiye-L, which includes aftershock data, with QMF-Türkiye-L1, which uses only the main earthquake event data. The inclusion of aftershock metadata resulted in a 5\% increase in precision, demonstrating its role in reducing false positives. However, other metrics such as AUC-ROC and F1 showed an improvement of 1\% and 2\% respectively. Additionally, we evaluated the QMF model (QMF-Türkiye-L-KAH) on building only located in the Kahramanmaraş region to further investigate the effect of incorporating aftershock metadata for the three strongest shaking events (M 7.8 in Pazarcık, M 7.5 in Elbistan and M 6.7 Nurdağı, all located in Kahramanmaraş Province). Relative to the baseline QMF-Türkiye-L1, QMF-Türkiye-L-KAH showed an additional 2\% increase in AUC-ROC and 3\% in F1 score. To assess how multiple earthquakes collectively influence damage prediction across regions, we plotted the Modified Mercalli Intensity (MI) for the five largest events and their cumulative MI (Figure \ref{fig:multiEQ}). The first three events (each $>=$ M 6.7) produced larger overlapping region of high‐intensity aftershocks in Kahramanmaraş region, leading to a pronounced combined effect on building damage. In contrast, the subsequent aftershocks ($<=$ M 6.7) generated minimal spatial overlap  and thus contributed little additional damage comparatively. This experimental analysis explains the importance of incorporating aftershock metadata and how varying degrees of aftershock overlap affect different regions.

\begin{figure*}
    \centering
    \includegraphics[width=1\linewidth]{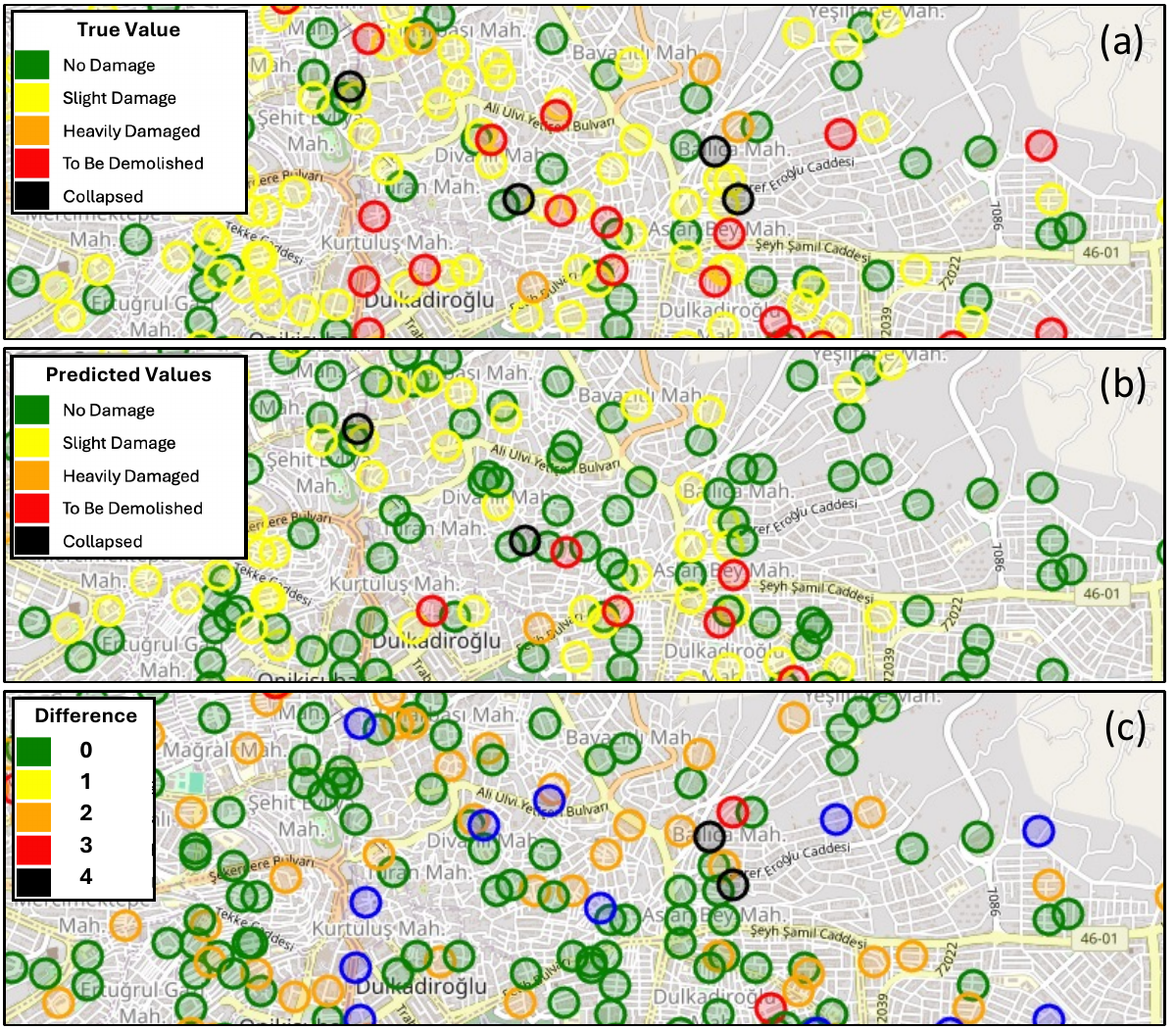}
    \caption{Evaluation Map for the city of Kahramanmaraş (QMF-Türkiye-S): (a) Ground Truth, (b) Predicted Values, and (c) Difference}
    \label{fig:Exp1_map}
\end{figure*}

We plotted the evaluation map for the city of Kahramanmaraş using the QMF-Türkiye-S model on the test dataset to assess the extent of misclassification (see Figure \ref{fig:Exp1_map}). The evaluation map is also useful for identifying the worst-hit regions after an earthquake. The map includes three key components: the true damage class of each building, the predicted damage class, and the absolute difference between the true and predicted values. The misclassification analysis reveals that most buildings are accurately classified, with 76.7\% of predictions being correct. Additionally, 17.7\% of the predictions were off by ±1 class, and 1.2\% by ±2 classes. Larger errors, with a difference of ±3 and ±4 classes, occurred in only 3.6\% and 0.8\% of cases, respectively. This analysis demonstrates the model's robustness in predicting building-level damage with very few significant misclassifications.

\subsection{Comparison of QuakeMetaFormer and Machine Learning Frameworks}

\begin{figure*}
    \centering
    \includegraphics[width=1\linewidth]{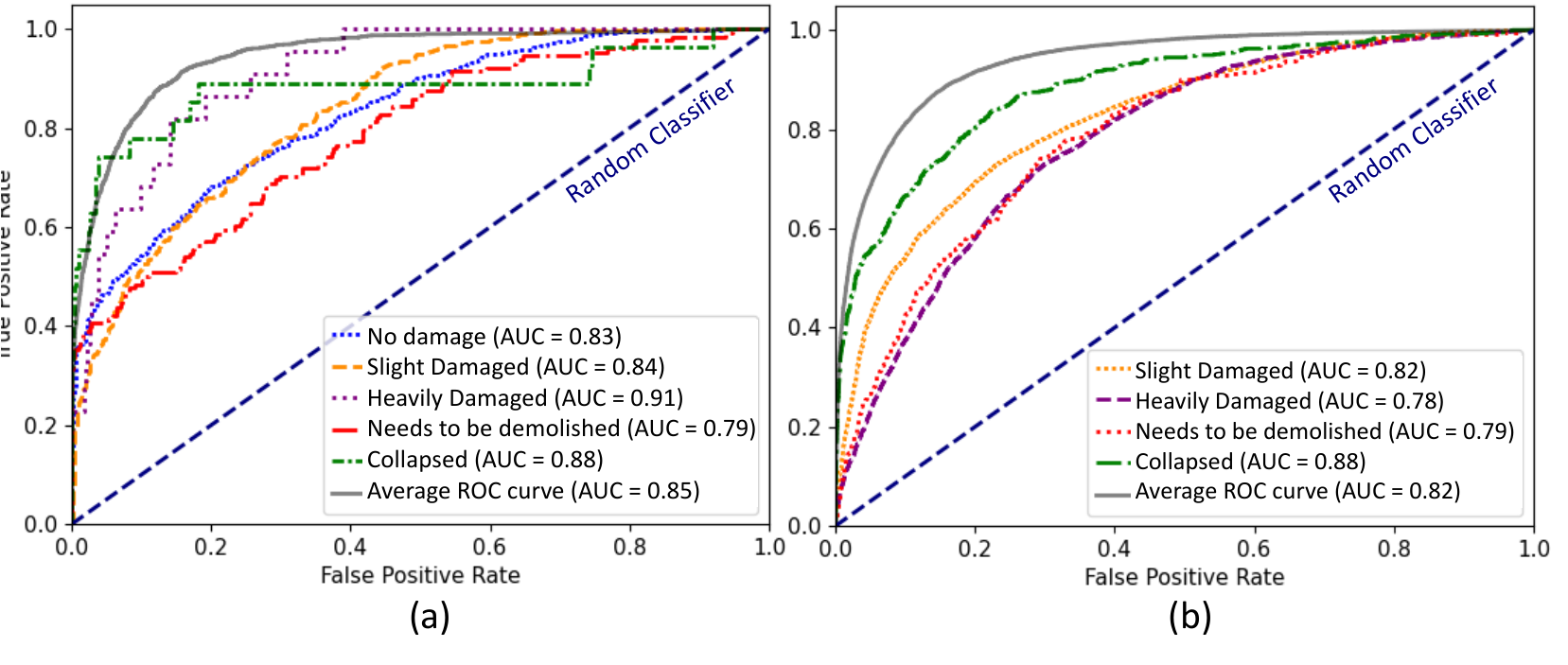}
    \caption{ROC curves for the (a) QMF-Türkiye-S and 
 (b) QMF-Türkiye-L}
    \label{fig:Exp2_roc}
\end{figure*}
We evaluate the effectiveness of our proposed models with state-of-the-art models published in the literature. \cite{YuIntelligentUpdates} proposed a machine learning approach using a random forest classifier trained on metadata from the Türkiye-S dataset. The model achieved an AUC-ROC of 0.69, outperforming DPM based methods by 11.25\%. We train our proposed QuakeMetaFormer with the entire Türkiye-S dataset including the satellite imagery and the results are shown in Table \ref{tab:experiment_summary_2_results}. Notably, QuakeMetaFormer achieves an AUC-ROC of 0.85, which is 16\% higher than the machine learning approach (RF-Yu). We also plot the ROC curve to understand the model's performance across each damage class (see Figure \ref{fig:Exp2_roc}a). The QuakeMetaFormer model demonstrated consistent performance across all classes, with a lowest AUC-ROC of 0.79 for class 3 (Needs to be demolished), indicating greater difficulty in distinguishing this class compared to others. A similar trend can be observed for QMF-Türkiye-L in Figure \ref{fig:Exp2_roc}b. It is important to note that a direct class-wise comparison of ROC scores between the two QuakeMetaFormer models is not appropriate, as they were trained on different datasets with varying numbers of classes, different set of buildings for training and validation, and varying input modalities. However, the results on both datasets highlights that QuakeMetaFormer architecture not only enhances overall model performance but also provides more balanced results across all damage classes, making it more reliable for building-level damage assessment in practice. 

Similarly, \cite{Rao2023EarthquakeLearning} evaluated their ensemble random forest framework for multiclass building damage classification across 4 earthquake events: the April 2015 Mw 7.8 Gorkha earthquake in Nepal, the September 2017 Mw 7.1 Puebla earthquake in Mexico, the January 2020 Mw 6.4 Puerto Rico earthquake, and the March 2020 Mw 5.3 Zagreb earthquake in Croatia. For multiclass classification, the reported F1 scores for these events were 0.34, 0.24, 0.23, and 0.33, respectively, with an average F1 score of 0.29. In contrast, the QuakeMetaFormer models achieved an average F1 score of 0.51 across the Türkiye-S and Türkiye-L datasets, demonstrating a significant improvement in multiclass classification (see Table \ref{tab:experiment_summary_2_results}). Although the models are not evaluated on the same event, making direct comparison challenging, the substantial difference in F1 scores (0.29 vs. 0.51) do highlights that existing machine learning methods face significant challenges in multiclass classification at the building level.

\begin{table}
\centering
\caption{Summary Table: QMF vs ML Frameworks}
\resizebox{0.47\textwidth}{!}{%
  \begin{tabular}{l c c c c c}
    \toprule
    \textbf{Experiment Name}     & \textbf{Source}            & \textbf{AUC-ROC} & \textbf{F1-Score} \\ 
    \midrule
    QMF-Türkiye-S                & Our Study     & 0.85 & 0.54 \\ 
    RF-Yu              & \cite{YuIntelligentUpdates}          & 0.69 & $-$ \\
    RF-Rao & \cite{Rao2023EarthquakeLearning} & $-$ & 0.29 \\
    \bottomrule
  \end{tabular}%
}
\label{tab:experiment_summary_2_results}
\end{table}

\subsection{Feature Importance Across Metadata} 

\begin{figure}
    \centering
    \includegraphics[width=1\linewidth]{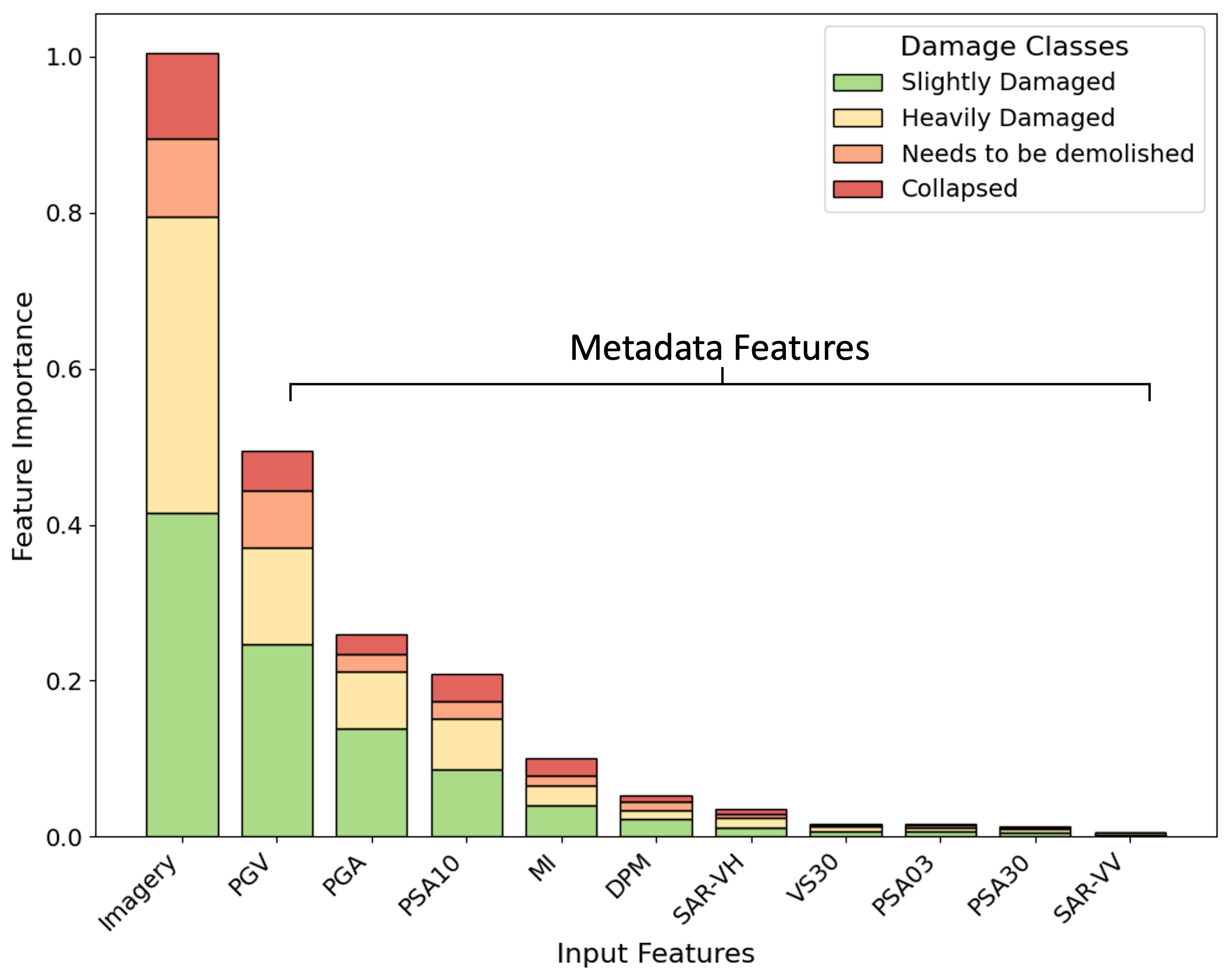}
    \caption{Model Input feature importance across damage classes}
    \label{fig:feature_importance}
\end{figure}

SHAP quantifies the contribution of metadata features and satellite imagery to model predictions and confidence in class assignments. Higher SHAP importance indicates stronger influence. However, higher value doesn’t always mean significant accuracy drop when a feature is removed, as other features can compensate. Figure \ref{fig:feature_importance} illustrates the impact of all input features, including imagery, on the model’s performance across four damage classes: Slightly Damaged, Heavily Damaged, Needs to be Demolished, and Collapsed. In the bar plot, as anticipated, satellite imagery demonstrates the highest contribution, as its detailed representation of visible building damage plays a crucial role in decision-making, far outweighing the influence of metadata.

Among metadata features, seismic intensity indicators such as PGV, PGA, and PSA show a substantial impact on predictions across all damage classes, highlighting their strong correlation with building damage. Notably, PSA values measured at 0.3 s, 1.0 s, and 3.0 s reveal that PSA at 1.0 s provides the most relevant information. This aligns with the natural frequency of many structures in the dataset, making it particularly relevant to structural response under seismic conditions. 

SAR-derived parameters exert a smaller influence on predictions. Within the group, the damage proxy is more informative than SLC amplitude SAR data, likely due to bitemporal data providing more insights into structural changes before and after the earthquake. Comparing  SLC amplitude SAR data, the results suggest that VH polarization contributes more to decision-making than VV polarization, indicating VH’s higher sensitivity to structural changes in buildings. VS30, which represents shear wave velocity, contributes little to the model; this may be due to its limited spatial variability at the building level, which reduces its effectiveness for multiclass damage classification. The limited contribution of SAR-derived parameters and VS30, compared to other features, may result from the low spatial resolution of these data sources, which fails to capture building-level changes necessary for accurate damage prediction.

\begin{figure}
    \centering
    \includegraphics[width=1\linewidth]{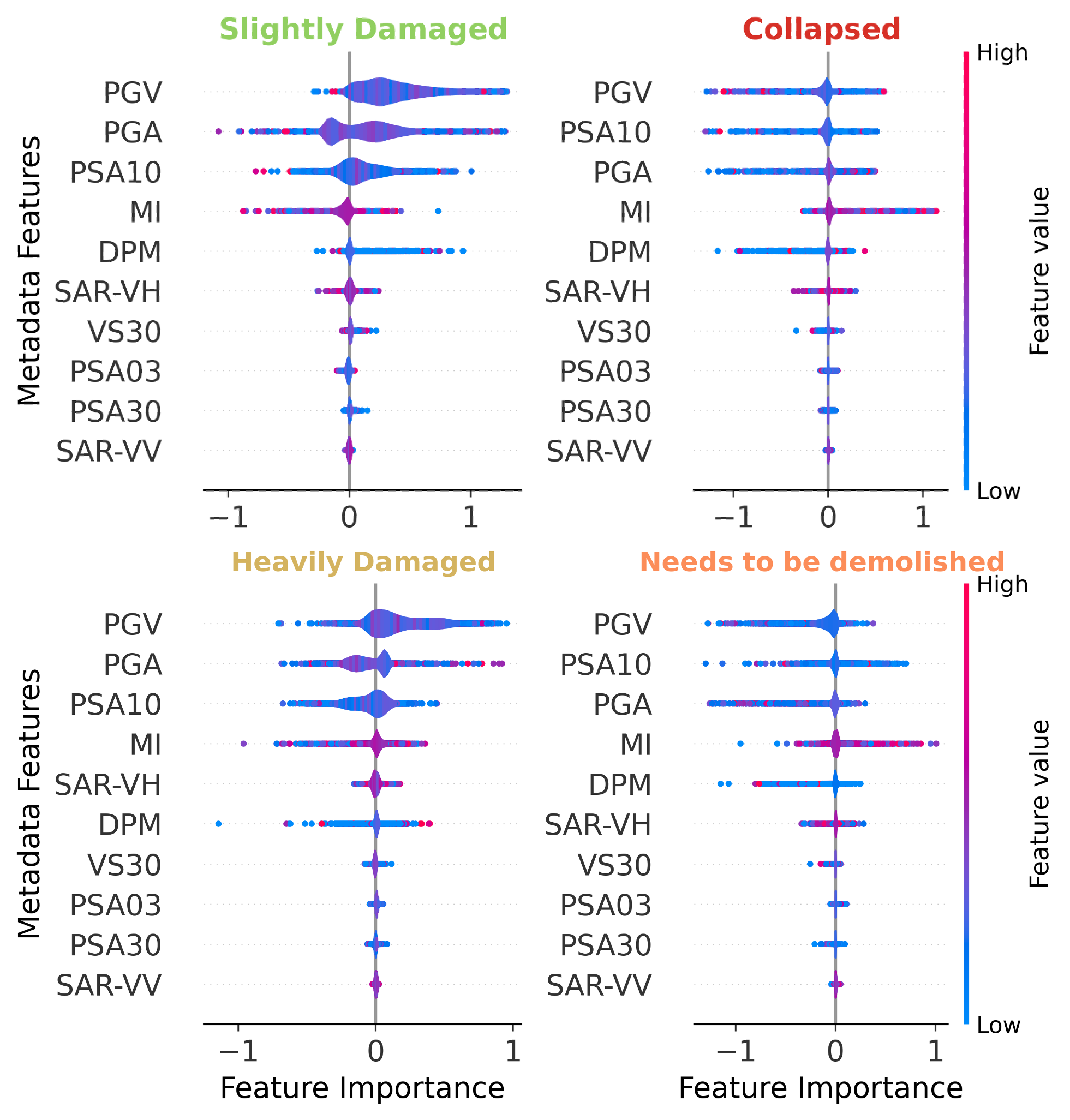}
    \caption{SHAP summary plots for Slightly Damaged (top left), Collapsed (top right), Heavily Damaged (bottom left), and  Needs to be demolished class (bottom right).}
    \label{fig:class0vs3}
\end{figure}

To understand the influence of metadata on each damage classes, we generated class-wise SHAP summary plots for all categories. Figure \ref{fig:class0vs3} consists of four subplots representing damage classes 1 through 4, highlighting the contributions of various features to the model’s predictions for each class.  The X-axis displays feature importance, showing how each feature influences the prediction, with values ranging from negative to positive. Positive values indicate that a feature supports the prediction towards the target class, while negative values suggest the feature pushes the prediction away from the specific class. Notably, negative values do not indicate that the feature negatively impacts the model's performance. Instead, they reflect whether the feature contributes to selecting the specific class or steers the prediction away from it. The Y-axis ranks metadata features by importance in descending order. The color gradient from blue to red represents each feature's value (low to high) within the test dataset, showing how high or low values of a feature affect the prediction. Each feature appear as a violin plot, showing the distribution of data points for that feature.

The SHAP summary plots reveal several key insights about feature contributions to damage class predictions. The contribution patterns are more distinct in the extreme damage classes, Slightly Damaged and Collapsed. For the top five features, low feature values support predictions toward the Slightly Damaged class. However, in the Collapsed damage class, these low feature values yield negative feature importance values, suggesting that they push predictions away from the class. Similarly, for metadata feature MI, high values (indicated in red) drive the model's prediction toward the Collapsed class and away from the Slightly Damaged class. This trend is also observed for the intermediate classes, Heavily Damaged and Needs to be Demolished, indicating that QuakeMetaFormer associates higher feature values with more severe damage (Classes 3 and 4) and lower values with less severe damage (Classes 1 and 2), thereby reducing misclassification by narrowing the gap between true and predicted classes. SAR-derived parameters (except DPM) and soil properties show minimal influence on predictions for both classes compared to other metadata features. Additionally, there are instances where high and low feature values intermingle and deviate from general trends, reflecting the complex relationships between features and damage class predictions. This complexity is particularly evident in intermediate classes. Overall, the plots illustrate how the QuakeMetaFormer model leverages metadata in its decision-making process, providing a detailed view of feature impacts on classification.

\subsection{Generalization Across Regions}

\begin{figure}
    \centering
    \includegraphics[width=1\linewidth]{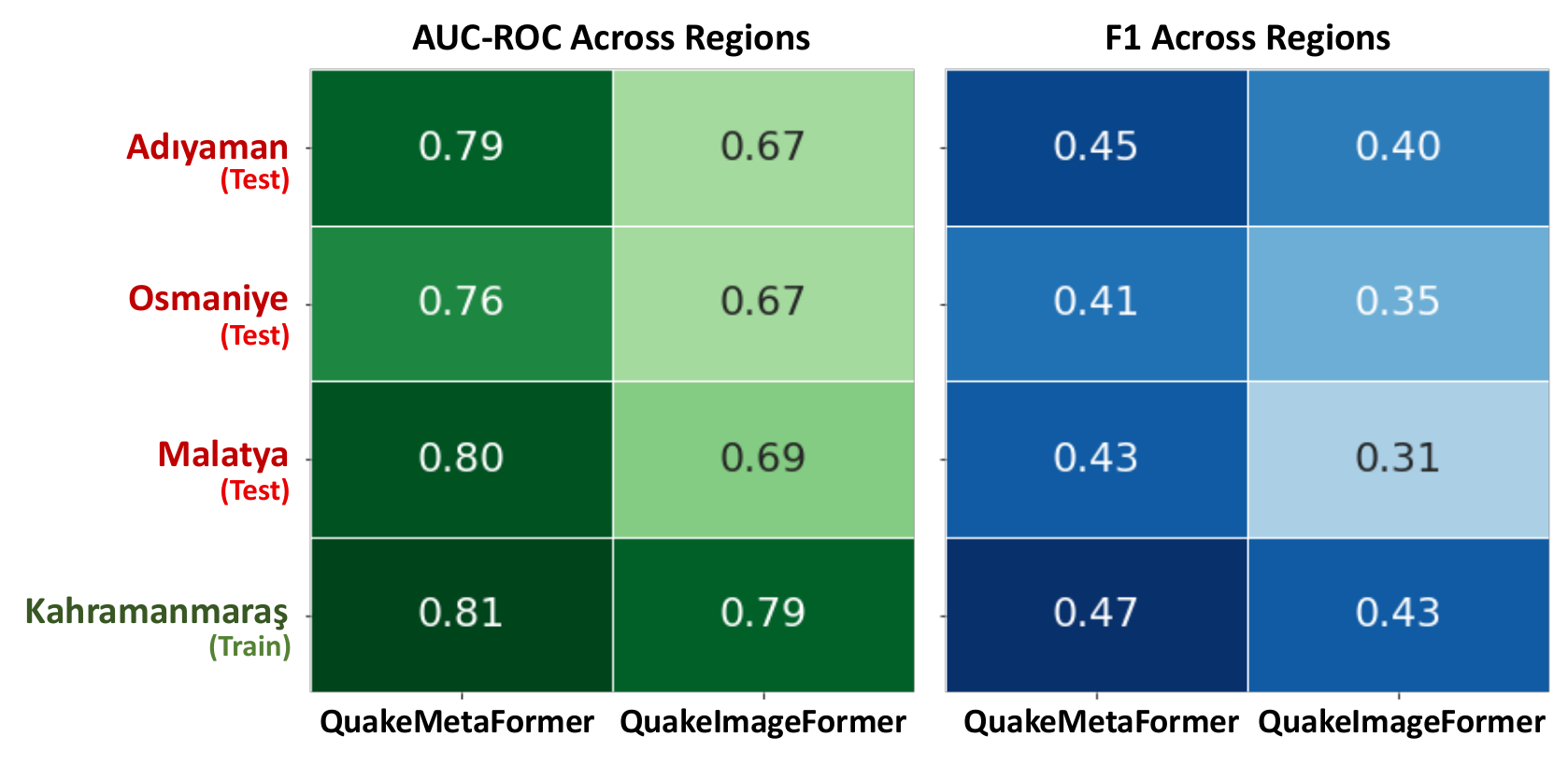}
    \caption{AUC-ROC (left) and F1 (right) scores across regions (y-axis) with models (QMF and QIF) on the x-axis.}
    \label{fig:Exp4_f1_roc}
\end{figure}

Figure \ref{fig:Exp4_f1_roc} summarizes the results of our experiments, aimed at evaluating the model’s ability to generalize across different regions affected by an earthquake event. The findings indicate that the model trained with both metadata and imagery (QMF) consistently achieved high AUC-ROC and F1 scores, demonstrating strong generalization to unseen regions. For instance, the QMF model trained on Kahramanmaraş region and tested on Adıyaman region, achieved an AUC-ROC of 0.79 and an F1 score of 0.45, while achieving 0.81 and 0.47, respectively, when both training and testing were conducted in Kahramanmaraş. This trend continues in Osmaniye and Malatya, where the QMF model demonstrates a stronger ability to generalize to unseen regions. We observe that testing the QMF model on unseen regions resulted in an average reduction (Train performance - Avg. performance on test) of only 0.027 in AUC-ROC and 0.04 in F1 score, highlighting the model’s strong generalization ability across different regions while maintaining performance.

We also trained the model without metadata (QIF) to assess the impact of adding metadata on generalization. As shown in Figure \ref{fig:Exp4_f1_roc}, the QIF model experienced a significantly larger drop in performance across unseen regions, with an average reduction of 0.11 in AUC-ROC and 0.08 in F1 score, compared to QMF model (with metadata). The results show an average relative improvement of 15.7\% in AUC-ROC and 21.8\% in F1 score when metadata is included (see Table \ref{tab:qmf_qif_comparison}). This difference highlights the positive impact of metadata on the model’s generalization ability to handle differences in data distribution between training and unseen testing regions, which is not as effectively achieved when using imagery alone, making it more robust and adaptable for practical applications. 

\begin{table}
\centering
\caption{Average performance of QMF and QIF trained on Kahramanmaraş and tested across Adıyaman, Osmaniye, and Malatya.}
\resizebox{0.47\textwidth}{!}{%
  \begin{tabular}{l c c}
    \toprule
    \textbf{Model/Metric}        & \textbf{AUC-ROC} & \textbf{F1-Score} \\ 
    \midrule
    QMF (imagery + metadata)          & 0.783            & 0.430             \\ 
    QIF (only imagery)            & 0.677            & 0.353             \\ 
    \midrule
    Difference (QMF - QIF)       & 0.106            & 0.077             \\ 
    Relative Improvement (\%)    & 15.7\%           & 21.8\%            \\ 
    \bottomrule
  \end{tabular}%
}
\label{tab:qmf_qif_comparison}
\end{table}

\section{Conclusion}
In this study, we proposed a novel framework for post-earthquake preliminary damage assessment (PDA), achieving state-of-the-art performance for the 5-class damage classification task. This problem has received limited attention in the literature because of the very challenging nature of making multiclass assessments after earthquakes. 

Our framework achieved an AUC-ROC of 0.85, surpassing the previous leading PDA framework by 16\%. Notably, the performance was consistent across all damage classes, with a minimum AUC-ROC of 0.79 for damage class 3 (Buildings that need to be demolished), and maximum AUC-ROC of 0.91 for damage class 2 (Buildings heavily damaged). We achieved this by addressing the key challenges of PDA by focusing on two critical components of the framework: diversifying data sources and enhancing data-processing approach. Instead of relying on a single modality, our proposed QuakeMetaFormer (QMF) integrates high-resolution post-earthquake satellite imagery with metadata, including ground motion, SAR, and soil data. To evaluate the framework, we conducted four experiments: analyzing the impact of metadata on model accuracy, comparing QuakeMetaFormer with other machine learning frameworks, assessing feature importance across metadata, and testing QuakeMetaFormer's generalization across regions. 

Our results show that adding metadata to satellite imagery enhances accuracy, yielding a 7\% improvement over the transformer model trained without metadata. Additionally, the model achieves an AUC-ROC of 0.8 even without high-latency metadata like SAR, demonstrating its robustness in scenarios with limited or no metadata. 

Compared to other ML-based methods, QuakeMetaFormer demonstrated superior performance in multiclass damage assessment, achieving an average F1-score of 0.51, significantly higher than the 0.29 achieved by competing approaches. We also conducted a detailed analysis to evaluate and understand the influence of individual metadata features on the model’s predictive performance, utilizing SHapley Additive exPlanations (SHAP). This analysis revealed that seismic indicators such as PGV, PGA, and SAR-derived parameters like DPM have a substantial impact across damage classes as compared to other parameters. 

Notably, incorporating metadata significantly enhanced the model's generalization capability, improving performance across different regions affected by the same earthquake event. The results demonstrated that adding metadata improved AUC-ROC by an average of 15.7\% and F1 score by 21.8\%. 

These improvements underscore the significant potential of our framework to accelerate disaster recovery efforts, including prioritizing inspection regions, streamlining the federal aid process, and estimating recovery costs.

\section{Future Work}

While the proposed PDA framework demonstrates state-of-the-art performance and generalization across different regions affected by an earthquake, two important challenges remain open for future research. First, generalization across distinct earthquake events, with varying building typologies, regional conditions, and seismic characteristic, has not yet been fully explored. Future work should validate the framework on multiple disaster datasets to assess robustness under diverse real-world scenarios. Lastly, the potential of low latency high-resolution SAR imagery should be further explored to enhance the accuracy and effectiveness of PDA frameworks.

\bibliographystyle{SageH}

\begin{acks}
The research was carried out at the University of Houston, under a contract with the National Aeronautics and Space Administration. The authors also acknowledge the use of the Carya Cluster and the advanced support from the Research Computing Data Core at the University of Houston to carry out the research presented here. 

\end{acks}

\begin{funding}
This work was performed at the University of Houston under a contract with the Commercial Smallsat Data Scientific Analysis Program of NASA (NNH22ZDA001N-CSDSA) and the NASA Decadal Survey Incubation Program: Science and Technology (NNH21ZDA001N-DSI). 

\end{funding}

\section{Data and Resources}
The satellite imagery used in this study were obtained through MAXAR Technologies' open data program \cite{TurkeyMaxar}. SAR data was accessed through the Alaska Satellite Facility (ASF) via NASA's Earthdata Search API \cite{EarthdataEarthdata}, while DPMs were extracted from the Jet Propulsion Laboratory's data repository \cite{JPLShare}. The seismic intensity indicators and VS30 data were obtained through the USGS website \cite{MSequence, Vs30Data}. The ground truth building damage data for Türkiye-S can be accessed via contacting corresponding authors \cite{YuIntelligentUpdates}. The Türkiye-L dataset is accessible at \cite{2023Online}, and password for the data can be obtained by contacting the authors of the data at info@yercizenler.org. The code for the architecture used in the study can be found at \url{https://github.com/deepankkumar/QuakeMetaFormer}.

\section{Declaration of conflicting interests}
The author(s) declared no potential conflicts of interest with respect to the research, authorship, and/or publication of this article.

\bibliography{references}

\end{document}